\documentclass[10pt,twocolumn,letterpaper]{article}

\usepackage{cvpr}              

\usepackage{graphicx}
\usepackage{amsmath}
\usepackage{amssymb}
\usepackage{booktabs}
\usepackage{arydshln}
\usepackage{textcomp}
\usepackage[accsupp]{axessibility}
\DeclareMathOperator*{\argmin}{arg\,min}

\usepackage[pagebackref,breaklinks,colorlinks]{hyperref}

\makeatletter
\@namedef{ver@everyshi.sty}{}
\makeatother
\usepackage{tikz}

\usepackage[capitalize]{cleveref}
\crefname{section}{Sec.}{Secs.}
\Crefname{section}{Section}{Sections}
\Crefname{table}{Table}{Tables}
\crefname{table}{Tab.}{Tabs.}

\def\cvprPaperID{6651} 
\def\confName{CVPR}
\def\confYear{2022}

\def\authornote#1#2#3{{\textcolor{#2}{{\small[#1: #3]}}}}
\iftrue
\newcommand{\cmira}[1]{\authornote{Mira}{red}{#1}} 
\newcommand{\cfangyin}[1]{\authornote{Fangyin}{blue}{#1}} 
\newcommand{\CL}[1]{\authornote{CL}{blue}{#1}} 
\newcommand{\MZ}[1]{\authornote{Michael}{red}{#1}}
\newcommand{\crohan}[1]{\authornote{Rohan}{magenta}{#1}} 
\else
\newcommand{\cmira}[1]{} 
\newcommand{\cfangyin}[1]{} 
\newcommand{\CL}[1]{} 
\newcommand{\MZ}[1]{}
\newcommand{\crohan}[1]{} 
\fi

\newcommand\blfootnote[1]{%
  \begingroup
  \renewcommand\thefootnote{}\footnote{#1}%
  \addtocounter{footnote}{-1}%
  \endgroup
}

\begin{document}

\title{Self-supervised Neural Articulated Shape and Appearance Models}

\author{
Fangyin Wei$^{1*}$\hspace*{-1pt}
\and
Rohan Chabra$^2$
\and
Lingni Ma$^2$
\and
Christoph Lassner$^2$
\and
Michael Zollhoefer$^2$
\and
Szymon Rusinkiewicz$^1$
\and
Chris Sweeney$^2$
\and
Richard Newcombe$^2$
\and
Mira Slavcheva$^2$\vspace*{2pt}
\and
$^1$Princeton University \qquad $^2$Reality Labs Research
}

\twocolumn[{
\renewcommand\twocolumn[1][]{#1}
\maketitle
\thispagestyle{empty}
\vspace{-1cm} 
  \def\iw{6.8em}
  \def\yoff{13.5ex}
  \def\xoff{-5ex}

  \begin{tikzpicture}[inner sep=0pt]

  \def\ix{3.9em}
  \node(pi0) at(0,0)                   {\includegraphics[width=\ix]{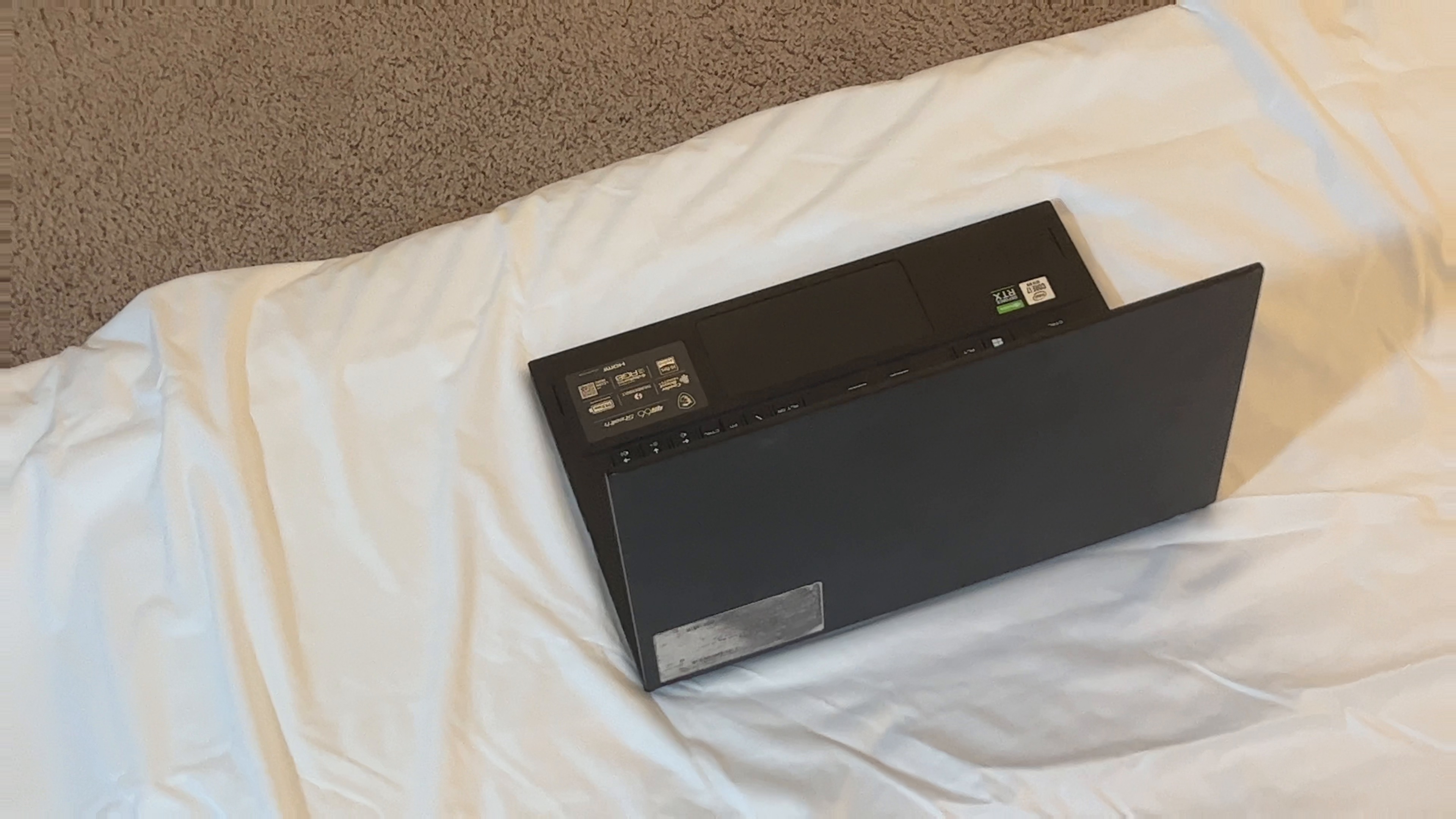}};
  \node(pi1) at(pi0.east)[anchor=west] {\includegraphics[width=\ix]{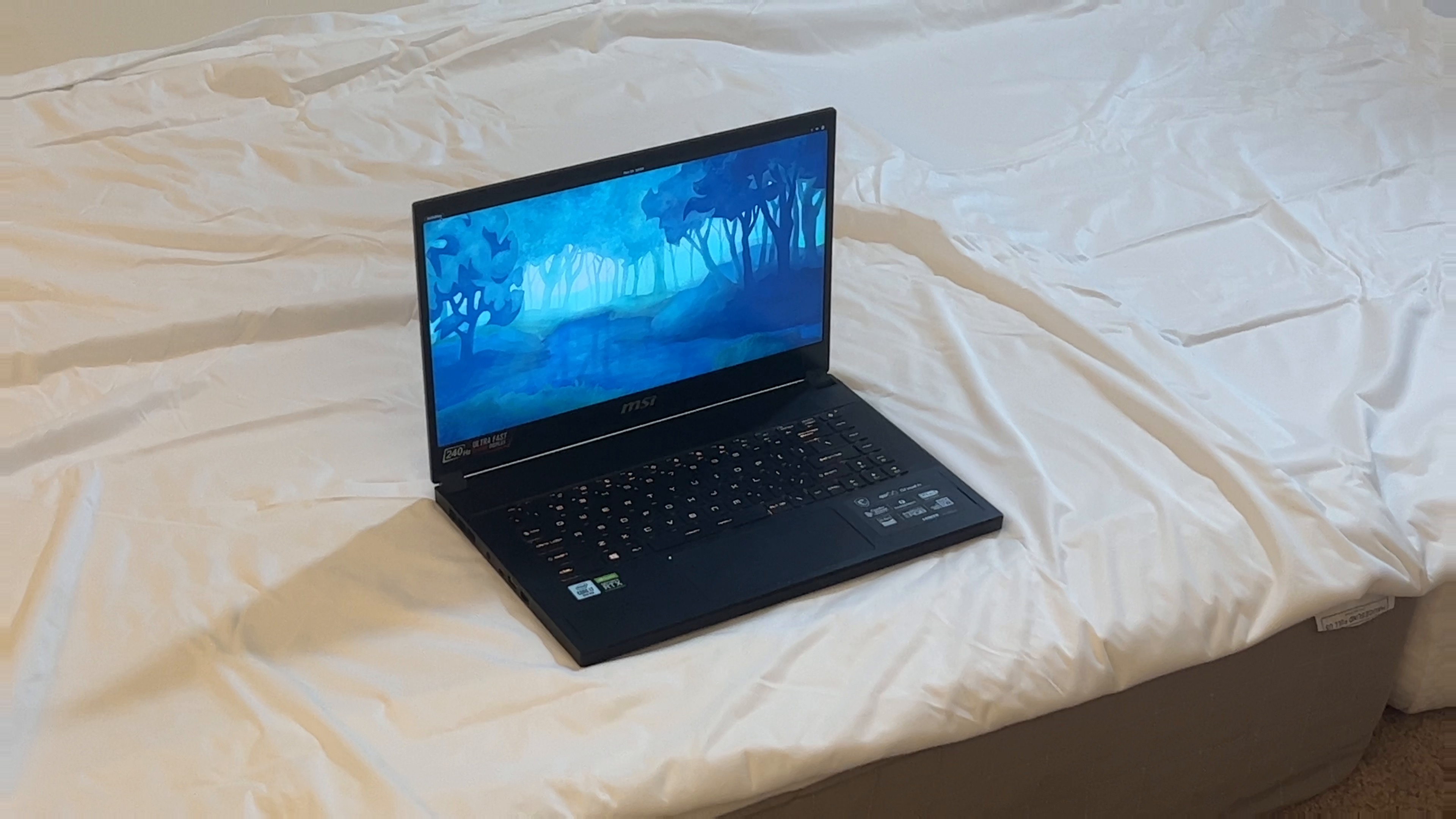}};
  \node(pi2) at(pi1.east)[anchor=west] {\includegraphics[width=\ix]{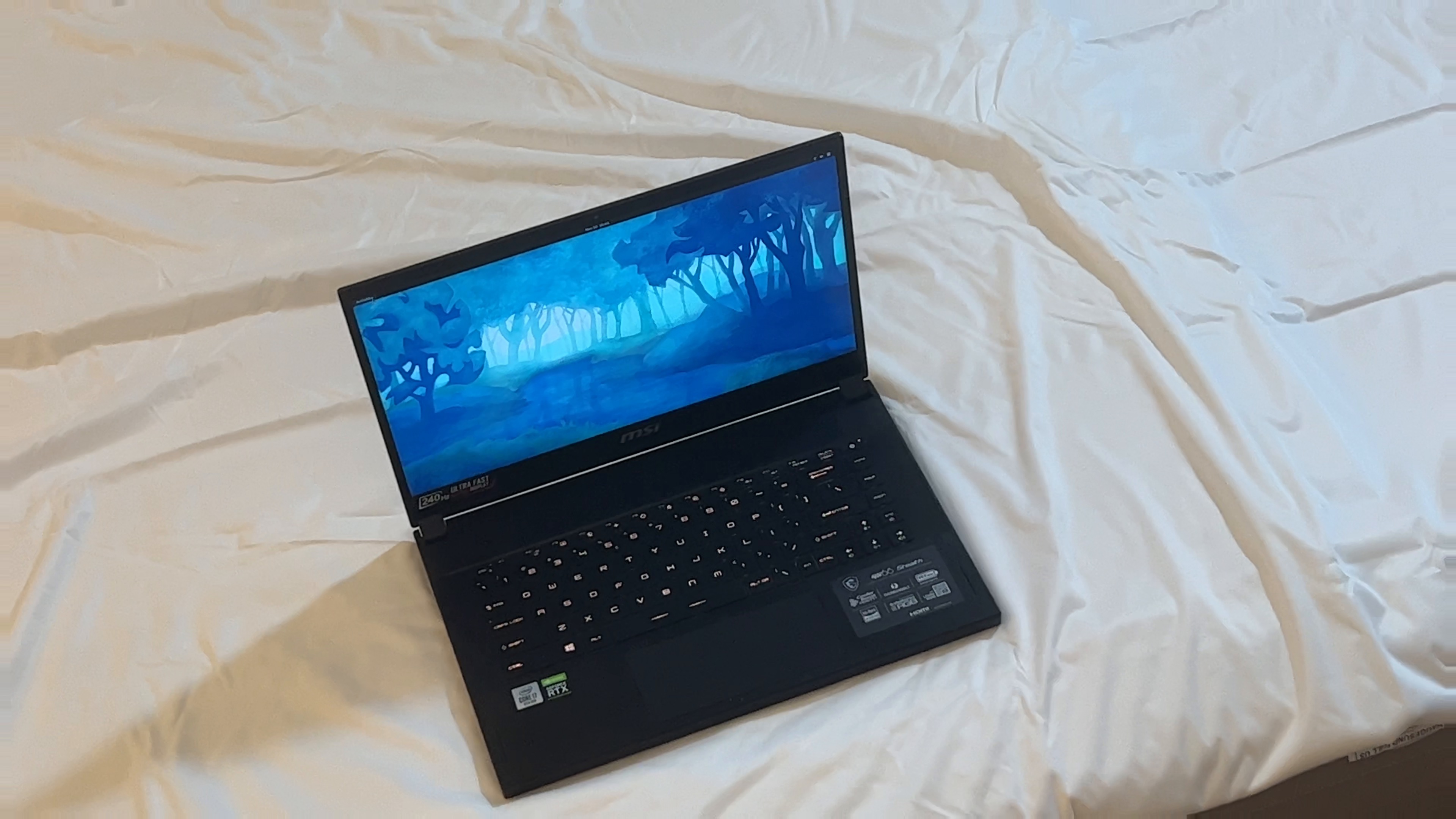}};
  \node(pi3) at(pi0.south)[anchor=north] {\includegraphics[width=\ix]{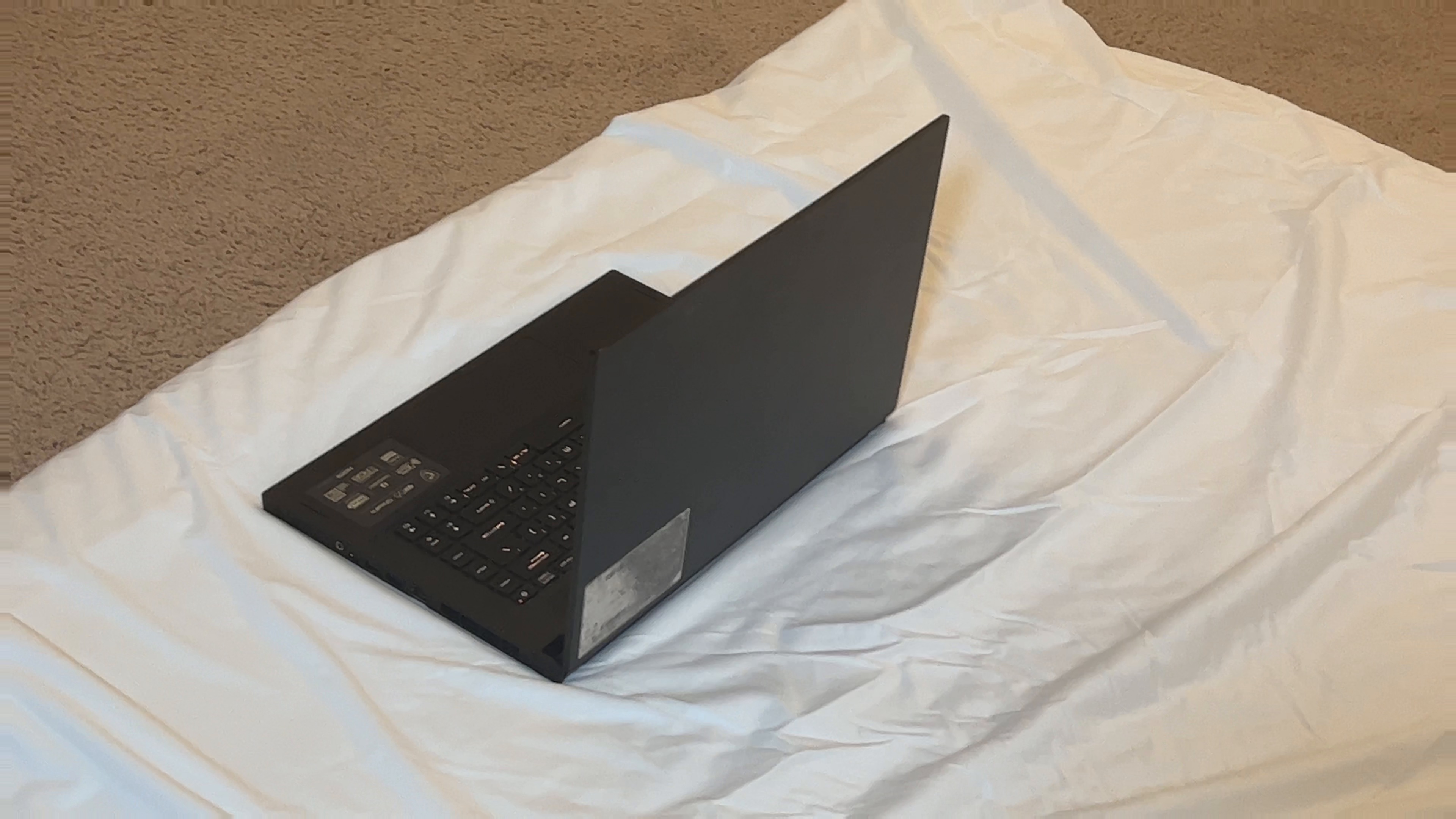}};
  \node(pi4) at(pi1.south)[anchor=north] {\includegraphics[width=\ix]{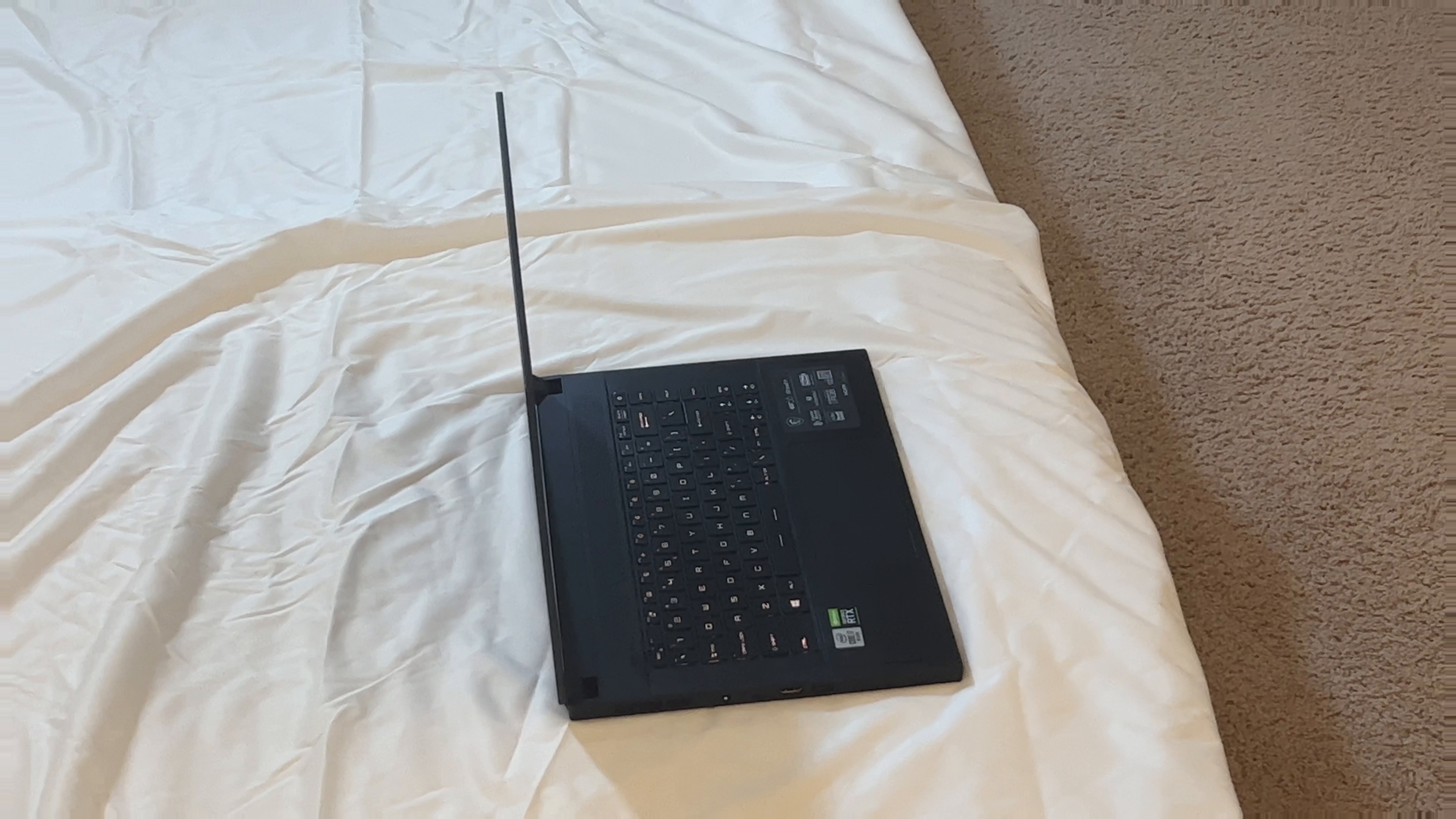}};
  \node(pi5) at(pi2.south)[anchor=north] {\includegraphics[width=\ix]{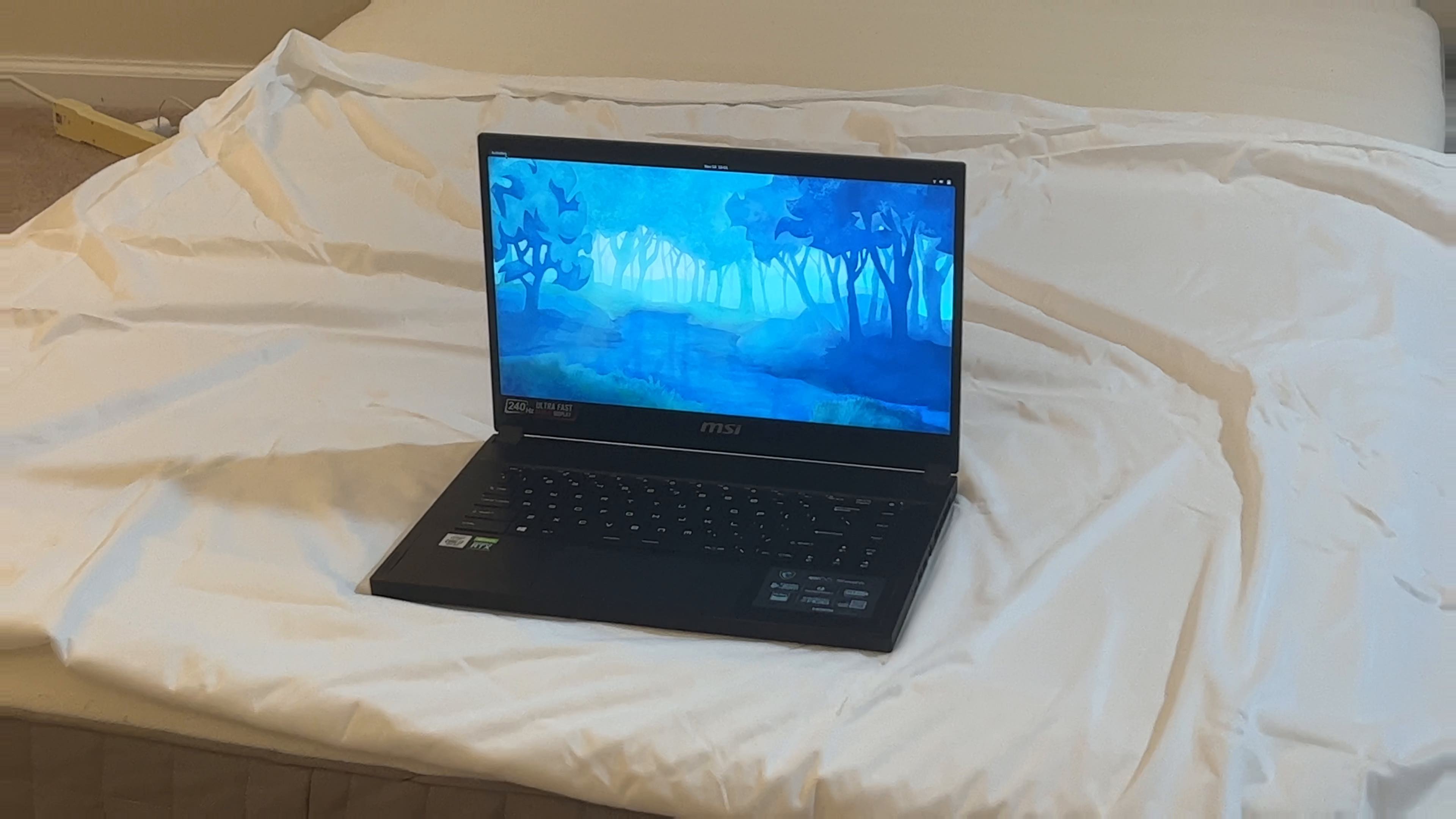}};

  \node(pr0) at(pi2.north east)[anchor=north west, xshift=1ex,yshift=1ex] {\includegraphics[width=6.5em]{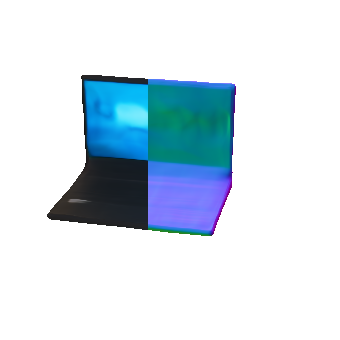}};

  \def\iw{5.2em}
  \node(p00) at(6.5,-0.2)                 {\includegraphics[width=\iw, ]{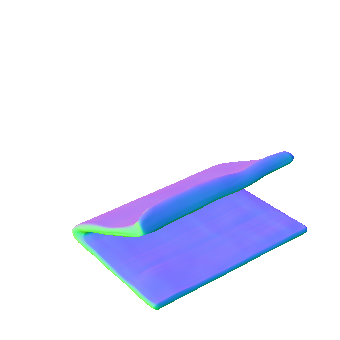}};
  \node(p01) at(p00.east)[anchor=west, xshift=-2ex] {\includegraphics[width=\iw]{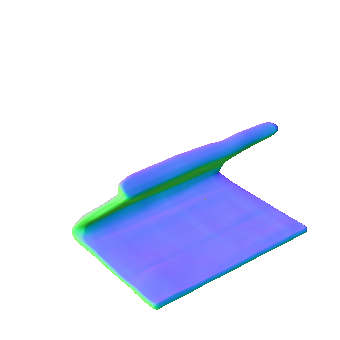}};
  \node(p02) at(p01.east)[anchor=west, xshift=-2ex] {\includegraphics[width=\iw]{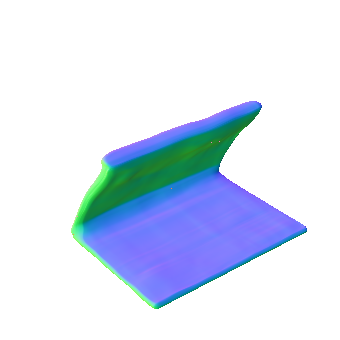}};
  \node(p03) at(p02.east)[anchor=west, xshift=-2ex] {\includegraphics[width=\iw]{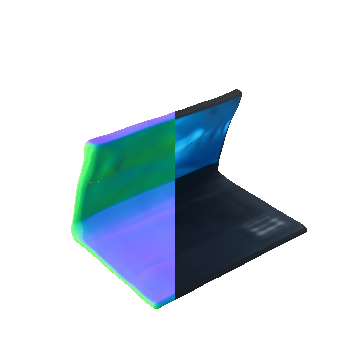}};
  \node(p04) at(p03.east)[anchor=west, xshift=-2ex] {\includegraphics[width=\iw]{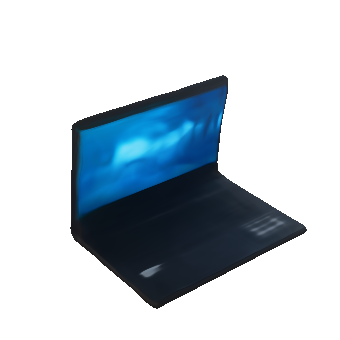}};
  \node(p05) at(p04.east)[anchor=west, xshift=-2ex] {\includegraphics[width=\iw]{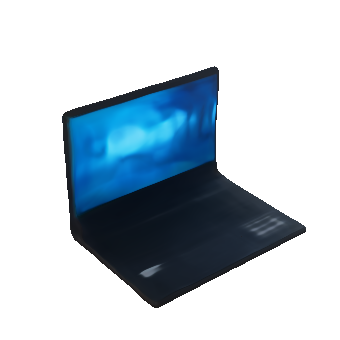}};
  \node(p06) at(p05.east)[anchor=west, xshift=-2ex] {\includegraphics[width=\iw]{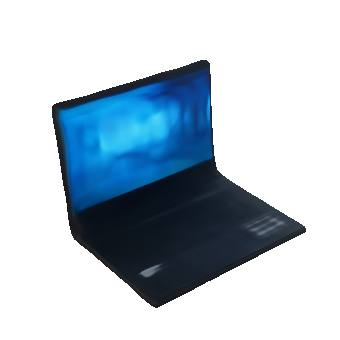}};


  \node(pj0) at(0, -1.7)                 {\includegraphics[width=\ix]{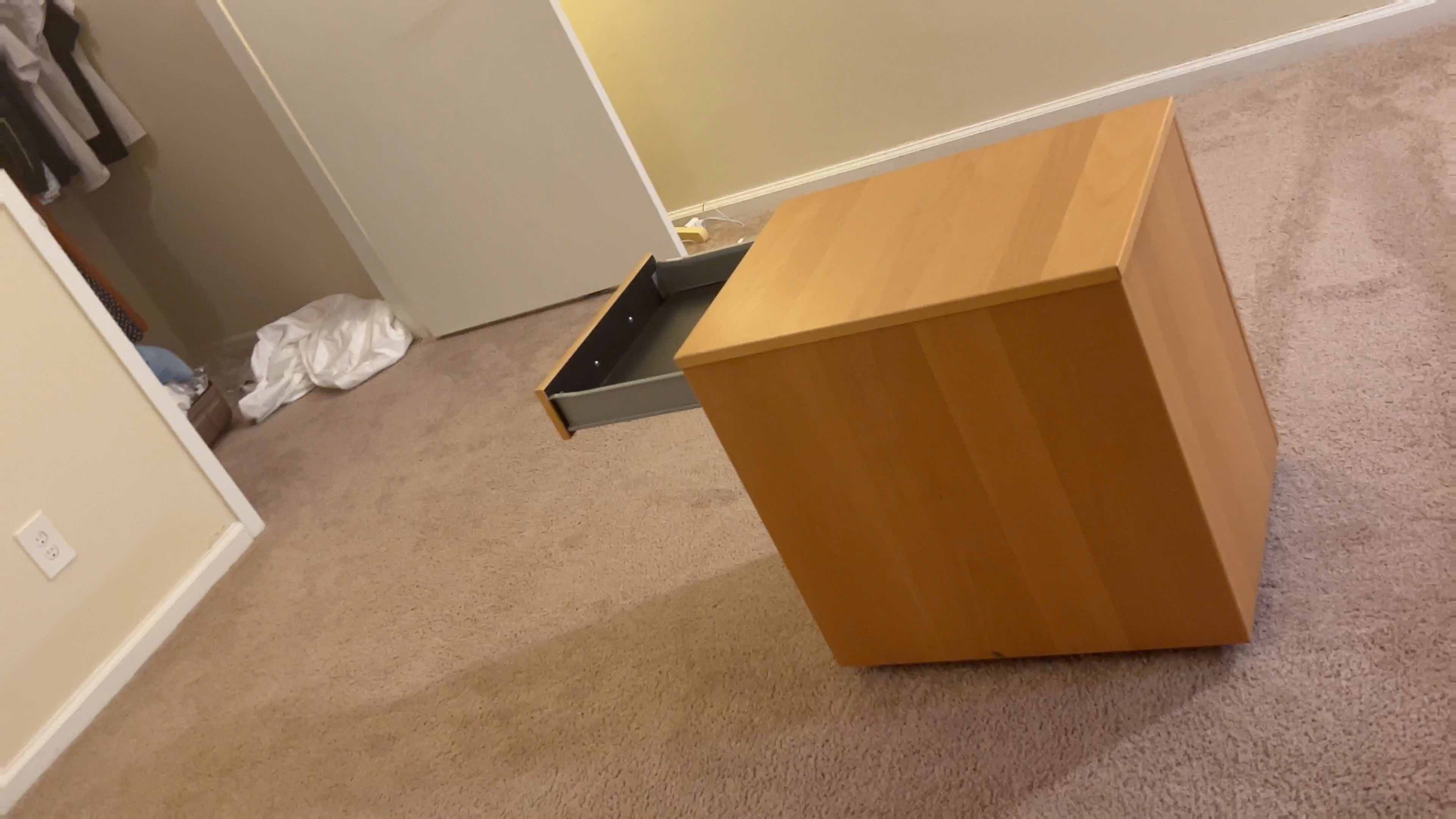}};
  \node(pj1) at(pj0.east)[anchor=west]   {\includegraphics[width=\ix]{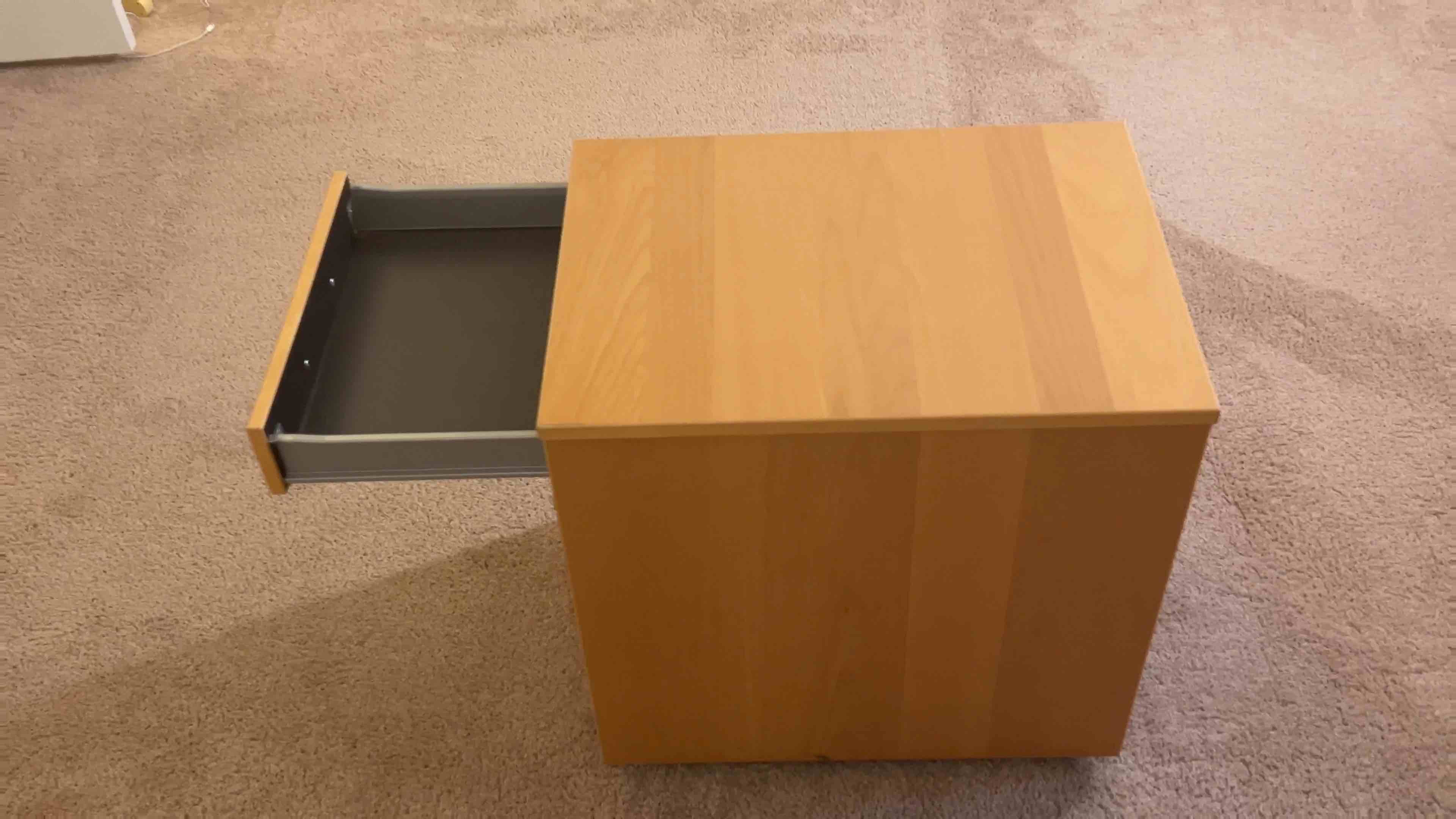}};
  \node(pj2) at(pj1.east)[anchor=west]   {\includegraphics[width=\ix]{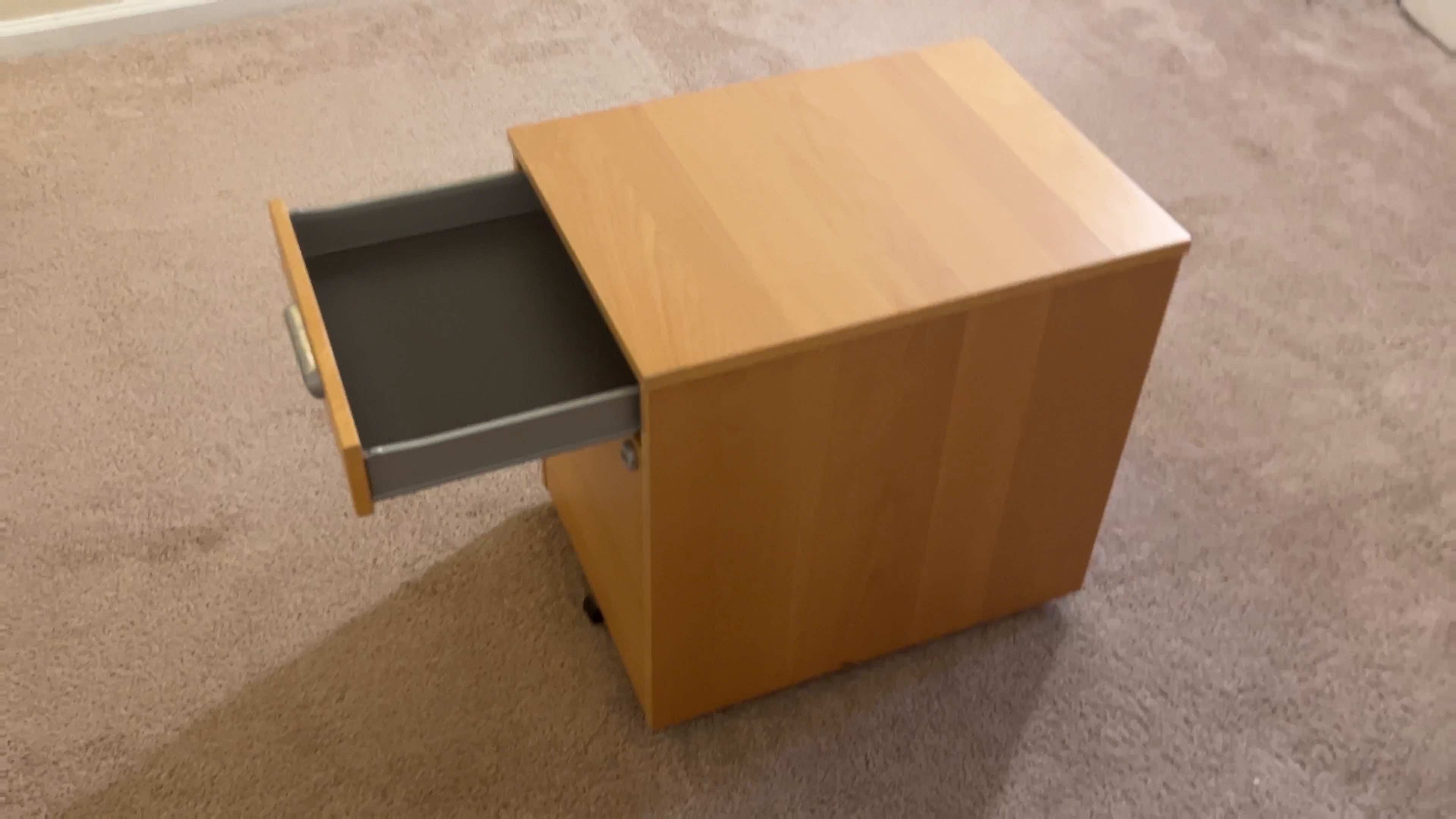}};
  \node(pj3) at(pj0.south)[anchor=north] {\includegraphics[width=\ix]{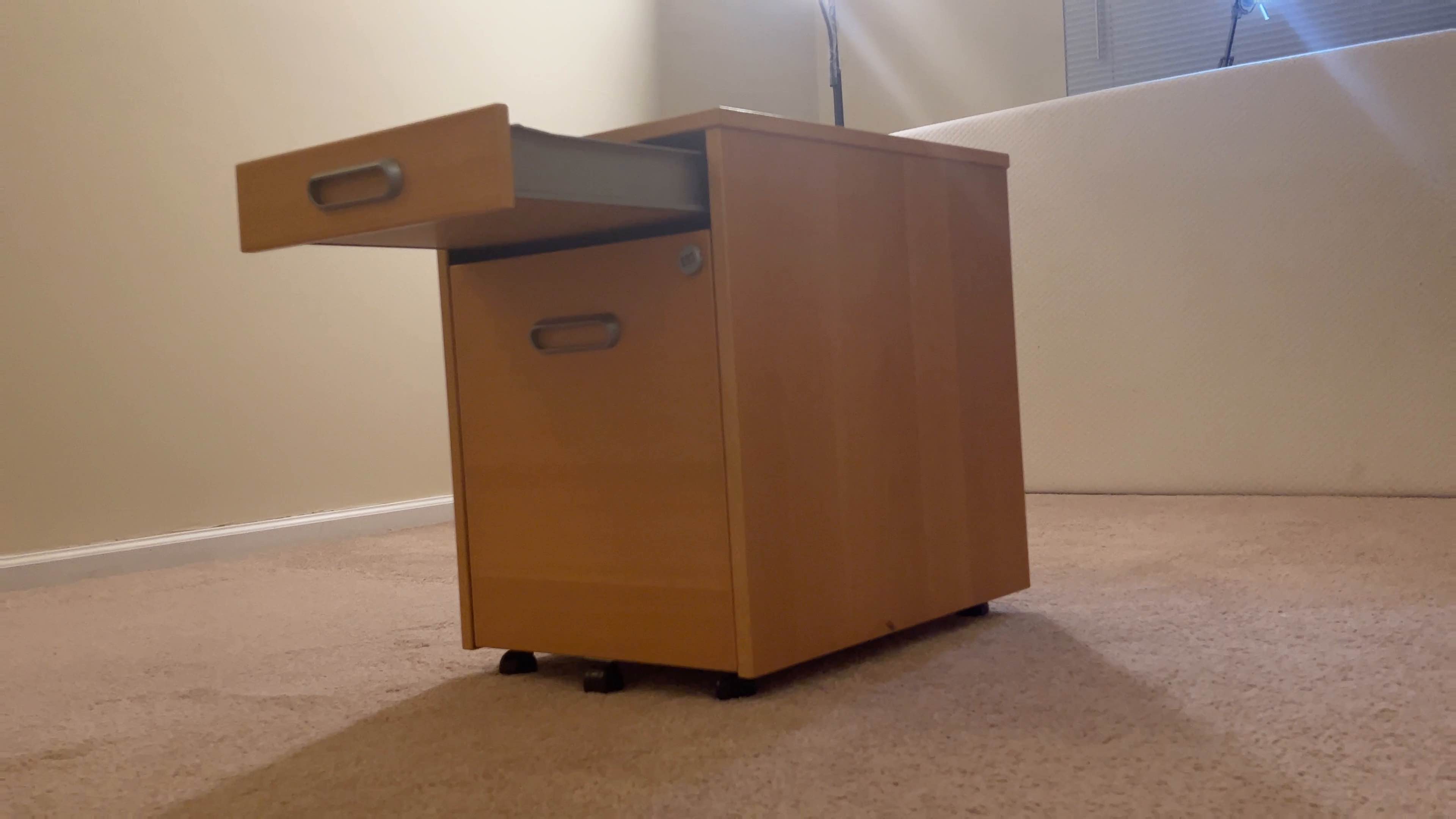}};
  \node(pj4) at(pj1.south)[anchor=north] {\includegraphics[width=\ix]{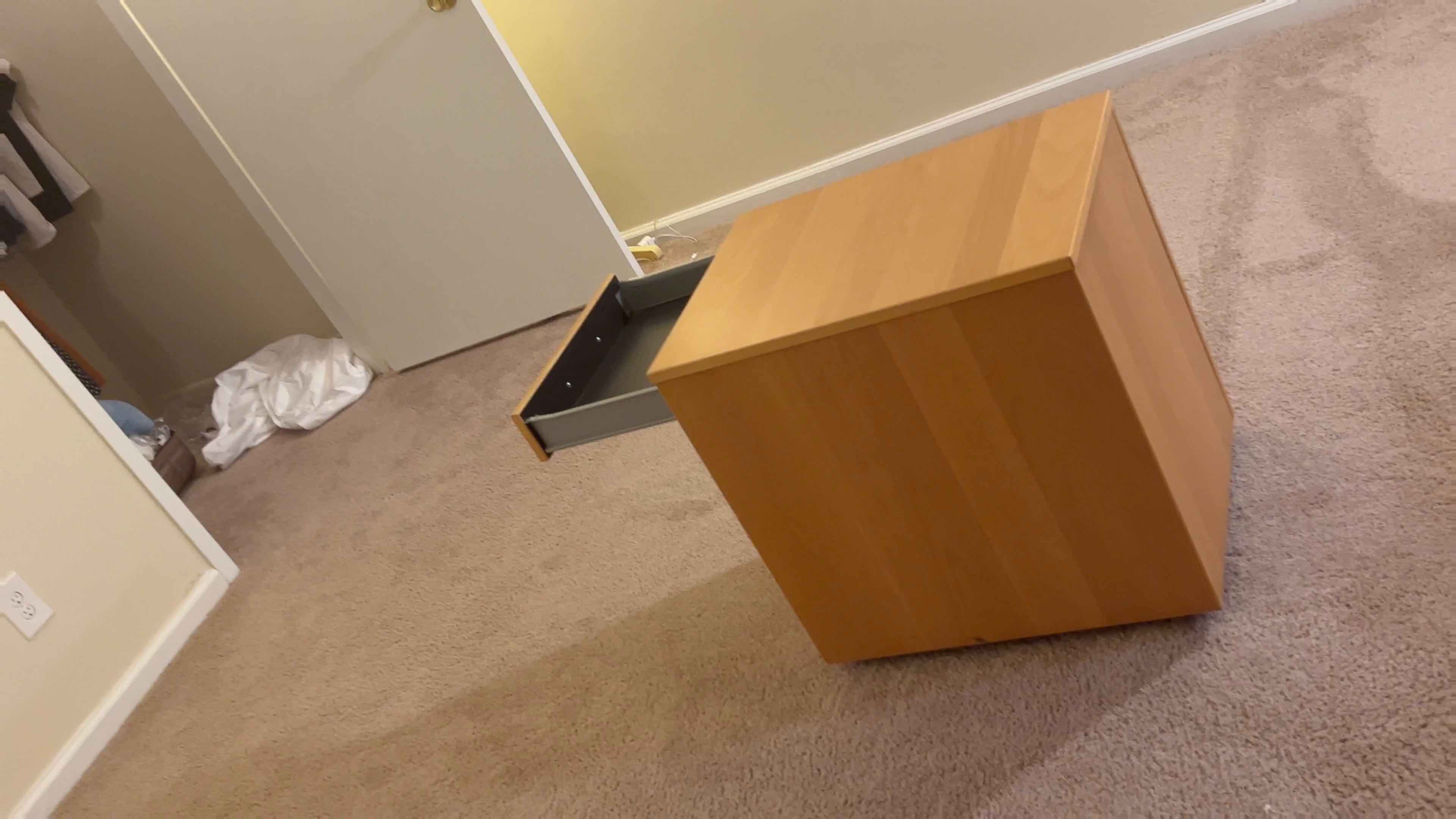}};
  \node(pj5) at(pj2.south)[anchor=north] {\includegraphics[width=\ix]{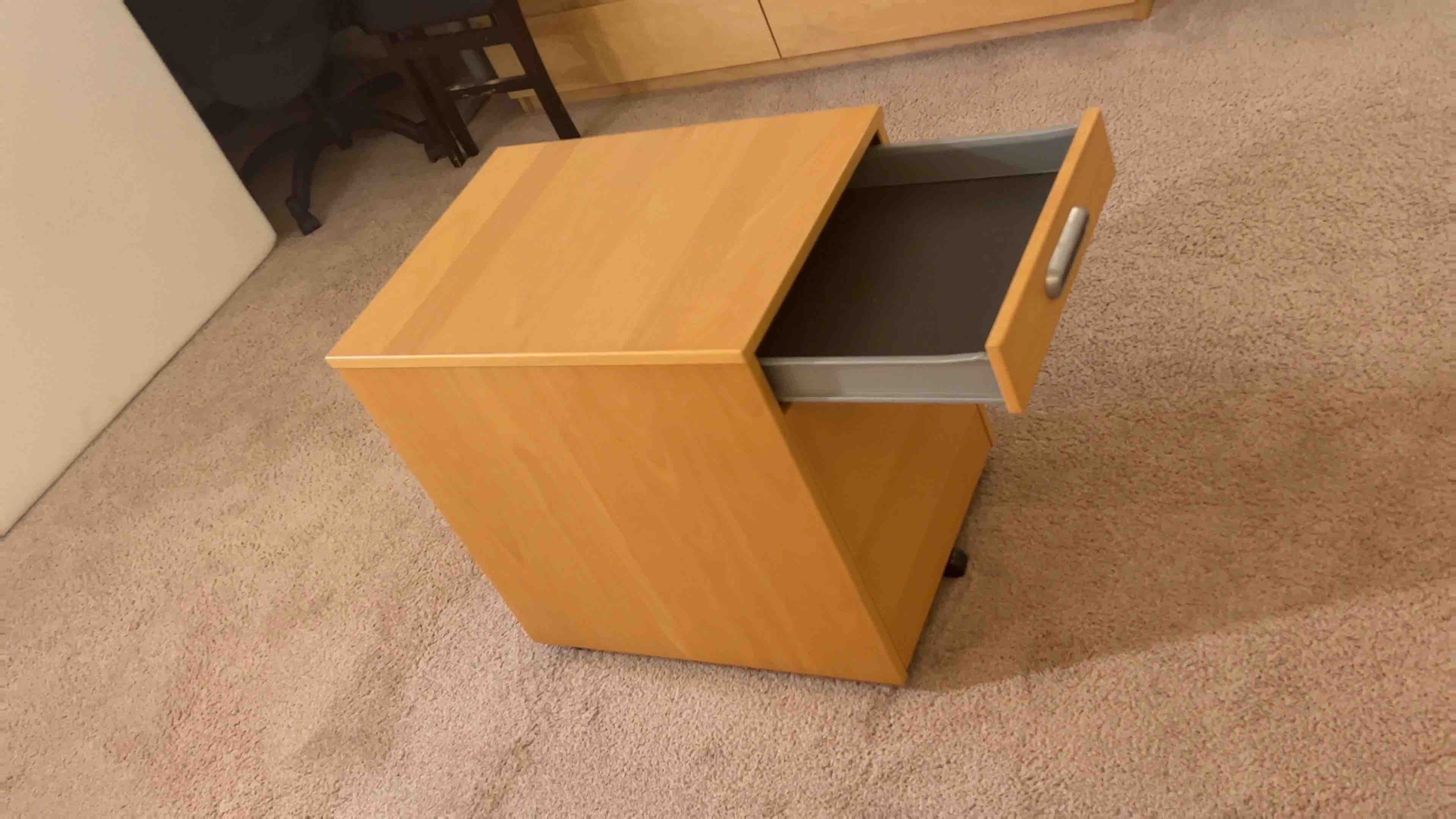}};
  
  \node(pr1) at(pj2.north east)[anchor=north west, xshift=-1ex,yshift=1ex] {\includegraphics[width=7.5em]{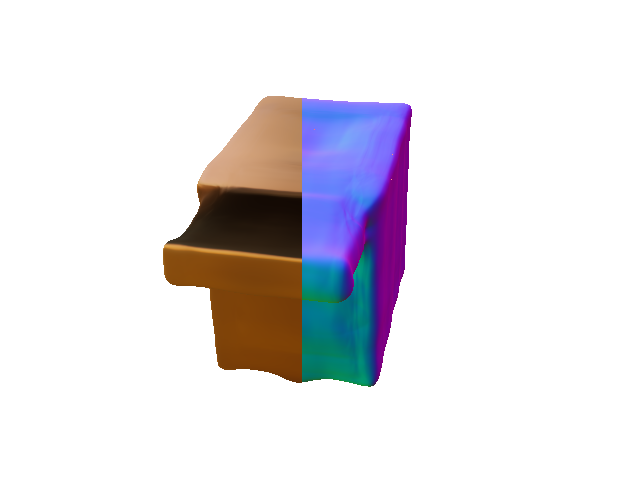}};

  \def\iw{7em}
  \node(p10) at(6.7,-2.1)                           {\includegraphics[width=\iw]{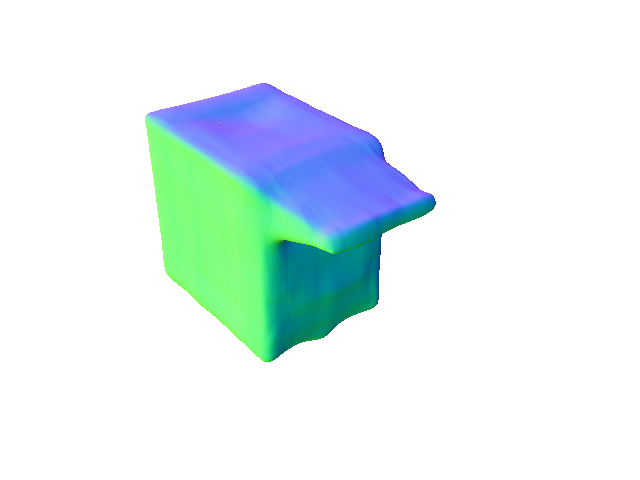}};
  \node(p11) at(p10.east)[anchor=west, xshift=-5ex] {\includegraphics[width=\iw]{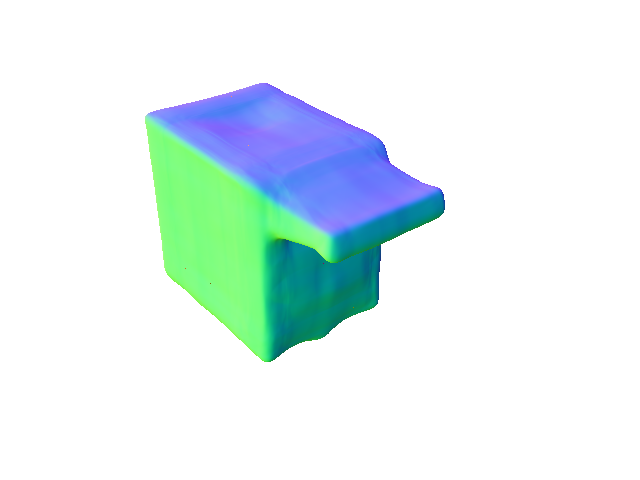}};
  \node(p12) at(p11.east)[anchor=west, xshift=-6ex] {\includegraphics[width=\iw]{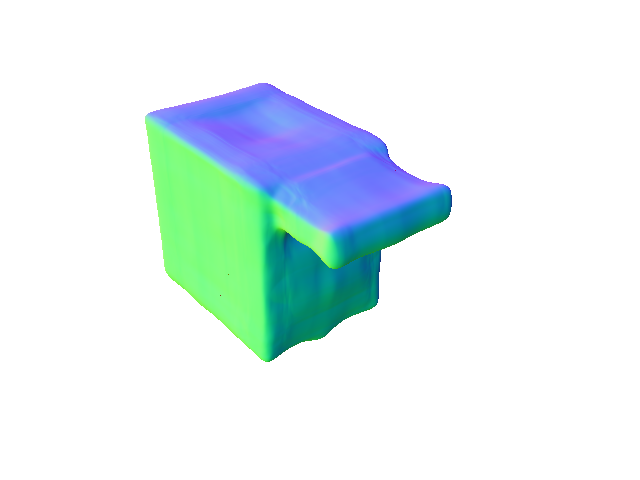}};
  \node(p13) at(p12.east)[anchor=west, xshift=-6ex] {\includegraphics[width=\iw]{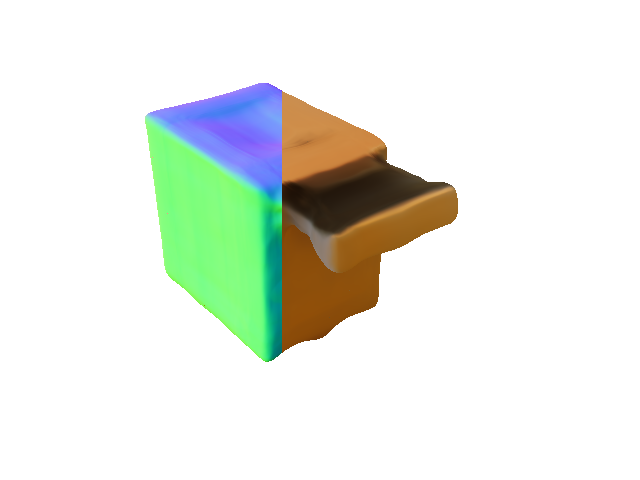}};
  \node(p14) at(p13.east)[anchor=west, xshift=-6ex] {\includegraphics[width=\iw]{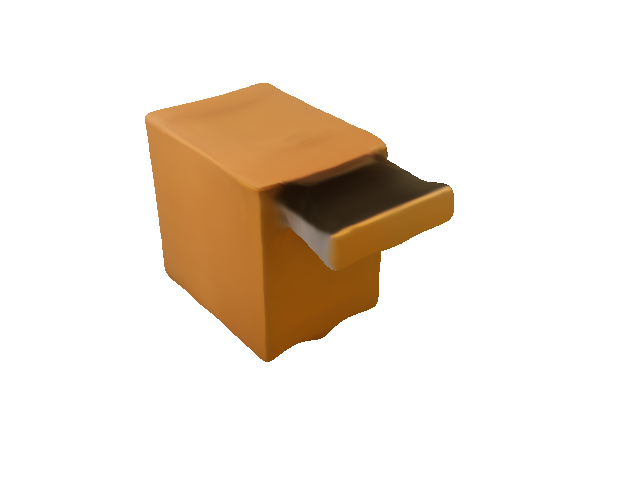}};
  \node(p15) at(p14.east)[anchor=west, xshift=-6ex] {\includegraphics[width=\iw]{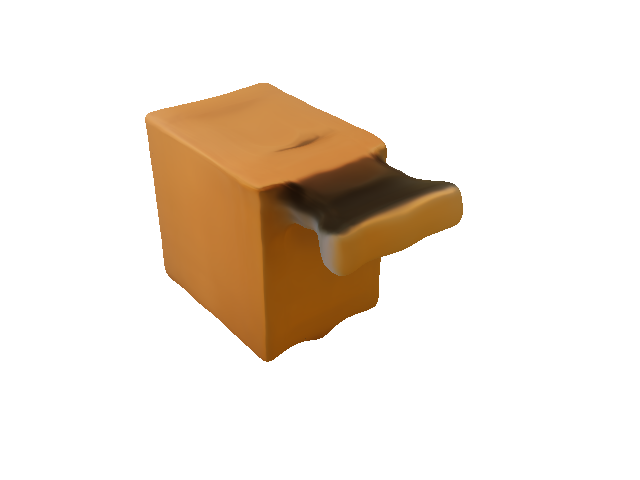}};
  \node(p16) at(p15.east)[anchor=west, xshift=-6ex] {\includegraphics[width=\iw]{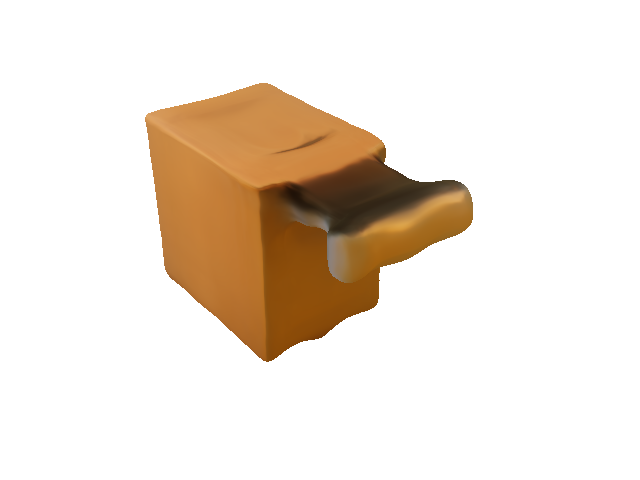}};
  
  \draw[->] (p10.south west)++(0.5,0.25) -- ++(10.5,0); 

  \end{tikzpicture}
  \vspace{-5ex}  
  \captionof{figure}{Our self-supervised method learns the shape and appearance of articulated object classes. After training from multi-view synthetic images of different states of object instances, our model can reconstruct and animate objects from static real-world images. \textit{Left}: input real views of a static object. \textit{Middle}: 3D reconstruction with shape and color. \textit{Right}: animation driven by the learned articulation codes.}
  \label{fig:teaser}
  \vspace{0.2cm}
}]
\maketitle

\blfootnote{$^*$Work begun during internship at Reality Labs Research.}

\begin{abstract}
%
Learning geometry, motion, and appearance priors of object classes is important for the solution of a large variety of computer vision problems.
While the majority of approaches has focused on static objects, dynamic objects, especially with controllable articulation, are less explored.
We propose a novel approach for learning a representation of the geometry, appearance, and motion of a class of articulated objects given only a set of color images as input.
In a self-supervised manner, our novel representation learns shape, appearance, and articulation codes that enable independent control of these semantic dimensions.

Our model is trained end-to-end without requiring any articulation annotations.
Experiments show that our approach performs well for different joint types, such as revolute and prismatic joints, as well as different combinations of these joints.
Compared to state of the art that uses direct 3D supervision and does not output appearance, we recover more faithful geometry and appearance from 2D observations only.
In addition, our representation enables a large variety of applications, such as few-shot reconstruction, the generation of novel articulations, and novel view-synthesis. Project page: \href{https://weify627.github.io/nasam/}{https://weify627.github.io/nasam/}.
\end{abstract}


\vspace{-0.7cm}

\section{Introduction}
\label{sec:intro}
Reconstructing articulated 3D objects from image observations in terms of their underlying geometry, kinematics, and appearance is one of the fundamental problems of computer vision with many important applications,~\eg in robotics and augmented/virtual reality.
This inverse graphics problem is highly challenging and of underconstrained nature, since image formation, \ie, mapping from the 3D world to discrete 2D pixel measurements, tightly entangles all visible properties of an object---and finding non-visible properties, such as kinematics, requires additional information such as information over time.

Most approaches that tackle this inverse problem rely on object/class-specific priors learned from large datasets with available 3D ground truth, which are challenging and expensive to collect.
The learned manifold of shape, appearance, and motion is often encoded via a low-dimensional latent space.
In the devised approaches, this learned low-dimensional prior is then used to better constrain the inverse reconstruction problem.

The majority of previous techniques in the literature has focused on reconstructing classes of static objects; dynamic objects, especially with controllable articulation, are less explored.
For example, occupancy networks condition the decision boundary of a neural classifier on a shape code to represent a class of static objects~\cite{Mescheder2019}.
Approaches such as DeepSDF follow a similar principle, but employ a learned continuous signed distance field (SDF)~\cite{Park2019} to model the object's surface implicitly as the zero level-set of a coordinate-based neural network.
DISN~\cite{Wang2019} further improves this technique and can recover more details.
All the mentioned approaches require dense 3D ground truth geometry for training and do not model object appearance.
One exception is IDR~\cite{yariv2020idr} which employs inverse differentiable rendering to reconstruct the shape (using an SDF) and view-dependent appearance of a single object, but this approach does not generalize to an entire object class.

Prior work on articulated deformations heavily focuses on humans and animals~\cite{smpl:loper:tog15,frank:joo:cvpr18, mano:romero:tog17, smal:zuffi:cvpr17, ghum:xu:cvpr20, nasa:deng:eccv20, leap:mihajlovic:cvpr21, Noguchi2021} due to the availability of large datasets and readily available (learned) priors.
One exception is the A-SDF~\cite{mu2021asdf} technique which is focused on general articulated objects.
It learns separate codes for shape and articulation and employs an SDF for representing the objects.
This approach learns a geometry prior across a class of articulated objects but does not jointly learn an appearance prior.
Additionally, it requires dense 3D ground truth for training.
We provide comparisons with A-SDF in our experiments.

Looking at these related works in context raises the question: 
Is it possible to jointly learn a prior over the 3D geometry, kinematics, and appearance over an entire class of articulated objects from only photometric 2D observations without requiring access to 3D ground truth?

We propose a novel approach for learning the geometry, kinematics, and appearance manifold of a class of articulated objects given only a set of color images as input.
Our novel 3D representation of articulated objects is learned in a self-supervised manner from only color observations without the need for explicit geometry supervision.
It enables independent control of the learned semantic dimensions.
Our model is trained end-to-end without requiring any articulation annotations.
Experiments show that our approach performs well on both of the most widespread joint types: revolute and prismatic joints, as well as combinations thereof.
We outperform the state of the art, even though these related approaches require access to ground truth geometry for explicit geometry supervision.
In addition, our approach handles a larger variety of joint types than A-SDF~\cite{mu2021asdf}.
Furthermore, our representation enables various applications, such as few-shot reconstruction, the generation of novel articulations, and novel view-synthesis.
In summary, our contributions are:
\begin{itemize}
    \setlength{\itemsep}{1pt}
    \setlength{\parskip}{1pt}
    \item A novel approach that learns a representation of the geometry, appearance, and kinematics of a class of articulated objects with only a set of color images as input.
    \item We introduce an embedding space for geometry, kinematics, and view-dependent appearance that enables a large variety of applications, such as the generation of new articulations and novel view synthesis.
    \item Our model, trained only on synthetic data, enables few-shot reconstruction of real-world articulated objects via fine-tuning, as shown in Fig.~\ref{fig:teaser}.
\end{itemize}

\section{Related Work}
\label{sec:relatedwork}
\paragraph{Articulated Object Modeling.}
In robotics, research on articulated objects focuses on kinematic models \cite{pillai:rss14, ben:corl19, screwnet:icra21}.
In the vision community, prior work on articulated deformations heavily focuses on humans and animals~\cite{smpl:loper:tog15,frank:joo:cvpr18, mano:romero:tog17, smal:zuffi:cvpr17, ghum:xu:cvpr20, nasa:deng:eccv20, leap:mihajlovic:cvpr21, Noguchi2021}.
Less explored is how to represent general objects with piece-wise rigidity.
Following the success of neural implicit representations, A-SDF~\cite{mu2021asdf} extends DeepSDF~\cite{Park2019} with separate shape and articulation codes to model category-level articulation.
By adding joint angles to the shape code, A-SDF learns a mapping to the corresponding deformed shape.
Instead of using known joint angles and dense 3D supervision as in A-SDF, we learn the articulation codes without the ground-truth labeling and only from images.
Li~et~al.,~\cite{naocs:cvpr20} propose the normalized articulated object coordinate space (NAOCS) as a canonical representation for category-level articulated objects.
This idea is further explored by CAPTRA~\cite{captra:iccv21} to track object articulation from point clouds, and adopted by StrobeNet~\cite{strobenet:arxiv21} to reconstruct articulated objects by first aggregating NAOCS predictions from multi-view color observations.
However, the reconstruction only provides geometry.
%
Recently, LASR~\cite{Yang2021} proposes a template-free approach to reconstructing articulated shapes from monocular video.
The algorithm jointly estimates the rest pose, skinning, articulation, and camera intrinsics by solving an inverse graphics problem resulting in coarse animatable meshes.

\vspace{-0.4cm} 
\paragraph{Deformation Fields for Shape Reconstruction.}
Shape deformation deals with deforming a shape to best fit a set of observations.
DynamicFusion~\cite{newcombe2015dynamicfusion} relied on local depth correspondences, follow-up methods used sparse SIFT features~\cite{innmann2016volumedeform}, dense color tracking~\cite{guo2017real} or dense SDF alignment~\cite{slavcheva2017killingfusion,slavcheva2018sobolevfusion}.
Such methods are subject to failure under fast motion due to the use of handcrafted functions for maintaining correspondences.
Recently, the performance of non-rigid tracking has been improved by data-driven approaches with learned correspondences~\cite{bozic2020deepdeform,bozic2021neural,li2020learning}.
Lately, there has been an exploration of neural generative models for shape deformation.
DIF~\cite{deng2021deformed} represents shapes by a template implicit field shared across the category, together with a 3D deformation field and a correction field dedicated to each shape instance.
%
%
FiG-NeRF~\cite{xie2021fig} approximates neural radiance fields of objects and simultaneously performs foreground/background separation.
NPMs~\cite{palafox2021npms} is a neural parametric model that learns latent shape and pose spaces to model 3D deformable shapes.
Unlike most of the above methods, we approximate geometry and view-dependent appearance of the shape. without any 3D supervision. 

\vspace{-0.4cm} 
\paragraph{Neural Representations for Geometry.}
Neural scene representations have been found compact and powerful to model the geometry and motion of objects~\cite{park2019deepsdf, Li2021}.
Implicit fields~\cite{Chen2019} use an implicit coordinate-based function and a latent code to model multiple object classes.
Local Deep Implicit Functions ~\cite{Genova2020} decompose space into a structured set of learned implicit functions to represent deformable shapes.
Occupancy networks~\cite{Mescheder2019} use a local neural classifier to represent objects.
Approaches such as DeepSDF~\cite{Park2019} follow a similar idea, but employ an SDF.
DISN~\cite{Wang2019} further improves this technique and can recover more details.
Local implicit grid representations~\cite{ChiyuMaxJiang2020, chabra2020deep} decompose a scene into local parts for which it learns implicit representations.
All of these methods only model the geometry of rigid objects.
\vspace{-0.4cm} 
\paragraph{Neural Representations for Appearance.}
Neural Radiance Fields (NeRF) model the appearance of static scenes via a coordinate-based scene representation~\cite{Mildenhall2020}.
Scene Representation Networks~\cite{Sitzmann2019} map world coordinates to a feature representation that can be translated to a rendering.
Differentiable Volumetric Rendering~\cite{Niemeyer2020} predicts a texture field.
Proxy-geometry, such as spheres or points can be used to speed up the rendering process~\cite{Trevithick2020,Ruckert2021}.
Point-based representations have been explored independently for view synthesis as well~\cite{Wiles2020,Aliev2020,Lassner2021}.
Neural Volumes breaks down the space into separate volumes with their own neural representations~\cite{Lombardi_2019_NeuralVolumes} and can model dynamic scenes.
For an in-depth discussion of recent neural rendering approaches, we refer to a recent survey \cite{tewari2021NeuralAdvances}.
\vspace{-0.4cm} 
\paragraph{Neural Representations for Motion.}
A hallmark feature of dynamic scenes is that they can be analyzed using flow fields, and these in turn can be used to represent them~\cite{li2021neural,du2021nerflow}.
There are several works that build on top of neural radiance fields to capture scenes in motion~\cite{xian2021space,Gao-freeviewvideo,pumarola2021d,Park2021,park2021hypernerf,attal2021torf,Li2021,Peng2021neuralbody}.
However, they do not allow to control the scene rendering.
Tretschk et al.~\cite{Tretschk2021}~allow to strengthen or weaken foreground motion.
All of the aforementioned works in this paragraph do not allow to control the resulting reconstruction.
Most prior work on controllable representations is around humans, for facial avatars~\cite{Gafni_2021_CVPR,Guo2021,Wang2020,Lombardi_2021_MVP} or human bodies~\cite{Noguchi2021,Su2021,Peng2021,liu2021neural,Xu2021,Huang2020}.
LASR~\cite{Yang2021} is a very general method that creates meshes for arbitrary objects.
It then provides a rigged model with an estimated skeleton to animate the object, however with very coarse results.
D-NeRF~\cite{pumarola2021d} approximates the radiance field of a deforming shape by using time as an input to the system but only works for a single scene.

\vspace{-0.2cm}
\section{Method}
\label{sec:method}
\begin{figure*}
    \centering
    \includegraphics[width=1\linewidth]{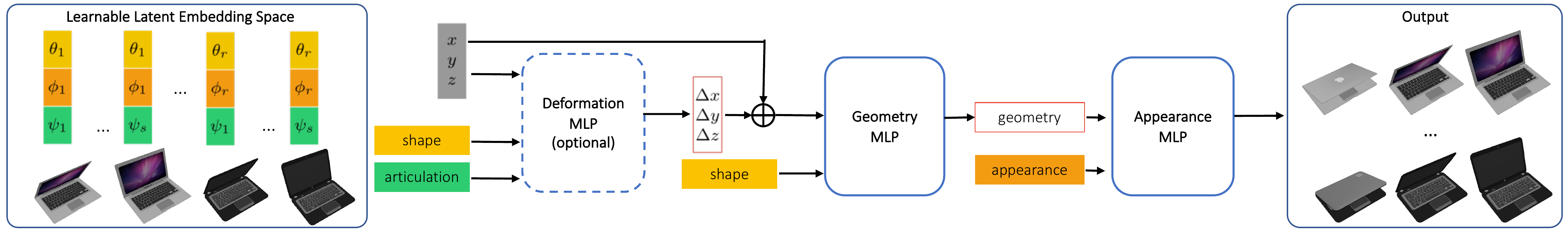}
    \vspace*{-0.5cm}
    \caption{\textbf{Framework overview.} Each object $i$ with articulation state $j$ is represented as $(\theta_{i}, \phi_{i}, \psi_{j})$, where each parameter encodes geometry, appearance, and articulation, respectively. A deformation network takes in the shape and articulation code in addition to query 3D locations (computed given camera parameters) to predict the displacement of the location. The displaced locations and the shape code pass through a geometry network to predict the geometry of the canonical pose shape. In the case of no deformation MLP, the geometry network directly takes in the 3D location, shape code, and articulation code. For the predicted geometry and given appearance code, an appearance network outputs an image according to input camera parameters.}
    \label{fig:overview}
\end{figure*}
Our goal is to build a representation that models the geometry and appearance of a class of articulated objects from RGB images without geometry priors. This representation must enable reconstructing unseen shapes and generating new articulations.
To this end, we utilize a differentiable rendering system with implicit
neural representations and learn a category-level embedding space with disentangled shape, appearance, and articulation representations. To enforce the disentanglement, different articulations of the same instance share the same shape code. The geometry is predicted by deforming a canonical shape conditioned on a learned articulation code. Without any joint annotations, the model is able to learn a continuous space for articulation that allows for generating new articulations.

We define an `articulated object' as an object which consists of several rigid parts connected by joints. We define further an \textit{articulation} of an articulated object as a specific state of its joints (\eg, a laptop with joint angles $40$\textdegree\ and $90$\textdegree\ are two distinct articulations). An object with a specific articulation is modeled by a representation set $(\theta, \phi, \psi)$ where $\theta\in\mathbb{R}^m, \phi\in\mathbb{R}^n, \psi\in\mathbb{R}^q$ are the code for geometry, appearance, and articulation, respectively.
In this section, we first review our backbone differentiable renderer and introduce a category-level embedding space (Sec. \ref{method:idr}).
Then we describe how to enforce disentanglement and the deformation field prediction (Sec. \ref{method:articulation} ). In Sec. \ref{method:training_inference}, we summarize the entire framework for training and inference.

\subsection{Differentiable Renderer with Category Priors}\label{method:idr}
We opt for Implicit Differentiable Renderer (IDR) \cite{yariv2020idr} as the backbone differentiable renderer.
While it originally works for a single object, we extend it to learn category-level geometry and appearance embedding. 
IDR is an end-to-end neural system that can learns 3D geometry, appearance, and camera extrinsics from masked 2D images and noisy camera pose initializations. There are three unknowns in IDR \cite{yariv2020idr}: geometry with learnable parameters $\Theta\in\mathbb{R}^r$,
appearance with learnable parameters $\Phi\in\mathbb{R}^s$, and camera extrinsics $\tau\in\mathbb{R}^k$. The geometry is represented as
\begin{equation}
    \mathcal{S}_\Theta=\{x\in \mathbb{R}^3 | f(x;\Theta)=0\},
\end{equation}
where $f$ models the signed distance function (SDF) to its zero level set $\mathcal{S}_\Theta$ (\ie the object's surface) \cite{yariv2020idr}. Given a pixel indexed by $p$, the rendered color of the pixel is
\begin{equation}
    L_p(\Theta, \Phi, \tau)=M(\hat{x}_p,\hat{n}_p,v_p; \Phi), 
\end{equation}
where $L_p$ is the surface light field radiance and $\hat{x}_p=\hat{x}_p(\Theta, \tau)$ denotes the first intersection of the ray $R_p$ and the surface $\mathcal{S}_\Theta$ with the corresponding surface normal $\hat{n}_p=\hat{n}_p(\Theta)$ and the viewing direction $v_p$. Both $f$ and $M$ are approximated by MLPs.

In the original IDR\cite{yariv2020idr}, each trained model only works for a single scene or object instance. We extend it to work across an entire class of objects by introducing additional embedding space. For each object instance $i$ from a class, we learn a geometry code $\theta_i\in\mathbb{R}^m$ and an appearance code $\phi_i\in\mathbb{R}^n$. During the learning process, all objects from the same category share the same geometry ($\Theta$) and appearance ($\Phi$) parameters.The new geometry and light field functions for object $i$ become:
\begin{align}
    \mathcal{S}_\Theta&=\{x\in \mathbb{R}^3 | f(x, \theta_i;\Theta)=0\}, \\
    L_p(\Theta, \Phi, \tau, \theta_i, \phi_i)&=M(\hat{x}_p,\hat{n}_p,v_p, \phi_i; \Phi).
    \label{lp}
\end{align}
 Leveraging the category-level prior embedded in $\Theta$ and $\Phi$ allows to reconstruct unseen objects from the same category by recovering only their geometry and appearance codes. 

\subsection{Code Sharing and Deformation Field}\label{method:articulation}
To encode the articulation, we further introduce an articulation code. For a category with $M$ training objects and $N$ sampled articulation states for each object, let shape $\mathcal{X}_{ij}$ denote an articulated object $i$ from a specific category with articulation state $j$. We jointly learn the representation $(\theta_{ij}, \phi_{ij}, \psi_{ij})$. Note that when articulations across different objects are aligned, we can further enforce disentanglement by having all objects from the same category share the same set of articulation codes and all articulation states of the same object share the same object code. Therefore, the representation for shape $\mathcal{X}_{ij}$ becomes $(\theta_{i}, \phi_{i}, \psi_{j})$, as illustrated in Fig. \ref{fig:overview}.

One observation is that geometry change happens both when articulating the same object and between different object identities. Learning to generate shapes by simultaneously processing the object identity and articulation state information may cause unwanted interference. Therefore, we split the shape prediction module into two parts: a geometry network $\mathcal{S}$ with parameters $\Theta\in\mathbb{R}^r$ and an (optional) deformation network  $\mathcal{D}$ with parameters $\Psi\in\mathbb{R}^t$. The former predicts the geometry of a canonical shape for each object given its shape code and is articulation-invariant. Conditional on a shape and articulation code that describes an articulated shape, the deformation network predicts a displacement of the query point to transform the query into the canonical space~\cite{Park2021,park2021hypernerf,Tretschk2021}. So the new geometry prediction flow is
\begin{align}
    x' = x + \mathcal{D}_\Psi&(x, \theta_i, \psi_j;\Psi), \\
    \mathcal{S}_\Theta&=\{x'\in \mathbb{R}^3 | f(x', \theta_i;\Theta)=0\}.
\end{align}
This separation of articulation prediction from the canonical shape helps further disentangle object identity and articulation state.

\subsection{Training and Inference}\label{method:training_inference}
During training, given camera intrinsic and extrinsic parameters and masked multi-view images, the model is trained to optimize both the latent embeddings $(\theta_{i}, \phi_{i}, \psi_{j})$ and network weights $(\Theta, \Phi)$. As shown in Fig. \ref{fig:overview}, each object $i$ with articulation state $j$ is represented as $(\theta_{i}, \phi_{i}, \psi_{j})$, where each parameter encodes geometry, appearance, and articulation, respectively. A deformation network takes in the shape and articulation code in addition to query 3D locations (computed given camera parameters) to predict the displacement of the location. The displaced locations and the shape code pass through a geometry network to predict the geometry of the canonical pose shape. For the predicted geometry and given appearance code, an appearance network outputs an image according to input camera parameters.
Let $I_p\in[0, 1]^3, O_p\in\{0,1\}$ be the RGB and mask values respectively for a pixel $p$ in an image taken with camera $c_p(\tau)$ and direction $v_p(\tau)$ ($p\in \Bar{P}$ indexes all pixels in the input collection of images), and $\tau$ represents the parameters of all the cameras in scene. The overall loss
function has the form:
\begin{multline}
 \mathcal{L}(\Theta, \Phi, \tau, \{\theta_{i}\}, \{\phi_{i}\}, \{\psi_{j}\})=\\L_{RGB} + \rho L_{mask}
    + \lambda L_E + \beta L_{code}.
    \label{total_loss}
\end{multline}

 We train on mini-batches of $P\subset\Bar{P}$ pixels sampled from one view of shape $\mathcal{X}_{ij}$ following IDR~\cite{yariv2020idr}.
The RGB loss is computed over regions where intersection has been found between the surface $\mathcal{S_{\Theta}}$ and Ray $R_p$ (\ie, $c_p+t_{p,0}v_p$ for $O_p=1$):
\begin{equation}
    L_{RGB}=\frac{1}{|P|}\sum_{O_p=1}|I_p-L_p(\Theta, \Phi, \tau, \theta_i, \phi_i,\psi_j)|,
\end{equation}
where $|\cdot|$ is $L_1$ norm and $L_p$ is defined in Eq.~\ref{lp}. The mask loss is
\begin{equation}
    L_{mask}=\frac{1}{\alpha|P|}\sum_{O_p=0}CE(O_p, S_{p,\alpha}(\Theta, \tau, \theta_i, \psi_j),
\end{equation}
where $CE$ is the cross-entropy loss and $S_{p,\alpha}=\text{sigmoid}(-\alpha \min\limits_{t\leq0}f(c+tv,\theta_i;\Theta)$  is an almost-everywhere-differentiable approximation to an indicator function for an object-occupied pixel $p$~\cite{yariv2020idr}. We enforce $f$ to be approximately a signed distance function with the Eikonal regularization following Implicit Geometric Regularization (IGR)~\cite{gropp2020implicit}: 
\begin{equation}
    L_E(\Theta)=\mathbb{E}_x(\|{\nabla}_x f(x, \theta_i, \psi_j;\Theta)\| -1)^2,
\end{equation}
where $x$ is distributed uniformly in a bounding box of the scene.
Lastly, following~\cite{park2019deepsdf}, we include a zero-mean multivariate Gaussian prior per latent code to facilitate learning a continuous shape manifold:
\begin{equation}
    L_{code}(\theta_i, \phi_i,\psi_j)=\|\theta_i\|_2^2 + \|\phi\|_2^2 + \| \psi_j\|_2^2.
\end{equation}

The goal during inference is to recover the representation $(\theta_i, \phi_i,\psi_j)$ given the RGB images of an unseen object. 
The three codes are randomly initialized, and then optimized through backpropagation with the following objective:
\begin{equation}
    \hat{\theta_i}, \hat{\phi_i},\hat{\psi_j}=\argmin_{\theta_i, \phi_i,\psi_j}\mathcal{L}(\Theta, \Phi, \tau, \theta_i, \phi_i,\psi_j).
\end{equation}
During optimization we can either fix the network weights (for in-distribution testing) or jointly optimize the network weights (for out-of-distribution testing).

\section{Experiments}
\label{sec:experiment}

\subsection{Experiment Setup}

\noindent\textbf{Dataset} 
%
All experiments use SAPIEN~\cite{xiang2020sapien}, a large-scale, public domain data set containing 2346 articulated objects across 46 categories.
We select six categories with representative articulation types and a sufficient number of instances: laptop, stapler, dishwasher, two-door fridge (LR for left and right, UD for up and down), eyeglasses, and storage furniture with drawer(s) (and door).
We use the SAPIEN simulation environment~\cite{xiang2020sapien} to render RGB images and corresponding masks.
During training and testing, we sample every 10\textdegree~ for rotational joints and 10 states in total for sliding joints. For multiple joints, we take all combinations of each single joint sampling. For each articulation, 60 views are sampled for training and 6 views for inference.
Please refer to supplementary materials for further details.

\noindent\textbf{Evaluation Metrics} 
We evaluate both the geometry and appearance of the predicted shape.
For geometry, we sample 30,000 points per shape~\cite{park2019deepsdf,mu2021asdf} and evaluate the Chamfer-L1 distance, which is the mean of the accuracy and the completeness score~\cite{Occupancy_Networks}.
For evaluating the rendered appearance, we report the Peak Signal-to-noise Ratio (PSNR).
All visualizations in this paper are rendered from unseen views. 

\noindent\textbf{Training and Inference}
For training, latent codes are randomly initialized with $\mathcal{N}(0,\frac{1}{l})$, where $l$ is the code length.
We set $\rho=100, \lambda=0.1, \beta=0.0001$ for the loss in Eq.~\ref{total_loss}.
We start with $\alpha=50$ and multiply it by a factor of 2 every 50,000 iterations (up to a total of 5 multiplications).
During inference, articulation codes are initialized to the mean of all learned articulation codes, while other codes are initialized as in training.
We run 600 iterations to recover the latent codes; if we do test-time adaptation~\cite{mu2021asdf}, we fine-tune both, model weights and codes, for another 600 iterations. 

\noindent\textbf{Baseline Methods}
The method variants that we compare are listed in Tab.~\ref{tab:methods}. 
For A-SDF~\cite{mu2021asdf}, we sample SDF values from SAPIEN data as described in~\cite{mu2021asdf} and run the author-provided code.
For IDR~\cite{yariv2020idr}, we use our own implementation in PyTorch which follows the original work, but without the global lighting feature: the SAPIEN dataset does not provide such effects and we empirically found this does not influence the result. 
Each method may additionally use test-time adaptation (TTA), as described in~\cite{mu2021asdf}, which in addition to optimizing the latent codes, optimizes the network weights during inference.

\subsection{Reconstruction}

\begin{table}[t]
\caption{\textbf{Methods used in experiments.} We specify whether they: handle static or articulated objects; share articulation code; use deformation field; output appearance or only geometry. We list the necessary input at train and test time. SDFs means SDF samples.}
\label{tab:methods}
\vspace*{-0.3cm}
\addtolength{\tabcolsep}{-1pt}
\centering
\resizebox{0.48\textwidth}{!}{
\begin{tabular}{lccccccc}
\specialrule{.2em}{.1em}{.1em}
Method & art./ & share & deform. & train & test & appear \\
name & static & art. & field & input & input & ance \\
\hline
A-SDF & art. & \checkmark & \texttimes & SDFs & SDFs & \texttimes \\
IDR 6 views & static & \texttimes&\texttimes & 6 RGBs & 6 RGBs & \checkmark \\
IDR 60 views & static & \texttimes&\texttimes & 60 RGBs & 6 RGBs & \checkmark \\
Ours-base & art. &\texttimes&\texttimes& 60 RGBs & 6 RGBs & \checkmark \\
Ours-Art & art. &\checkmark& \texttimes& 60 RGBs & 6 RGBs & \checkmark \\
Ours-Def & art. & \texttimes& \checkmark& 60 RGBs & 6 RGBs & \checkmark \\
Ours-ArtDef & art. &  \checkmark&\checkmark & 60 RGBs & 6 RGBs & \checkmark \\
\specialrule{.1em}{.05em}{.05em}
\end{tabular}
}
\end{table}

\begin{table*}[t]

\caption{\textbf{Comparison for reconstructing unseen synthetic shapes (Chamfer-L1).} We compare our method with A-SDF~\cite{mu2021asdf} and IDR~\cite{yariv2020idr}. As training IDR for each object is too computationally expensive, we present the average across 2 articulation states for 2 objects from each category. IDR is tested on the same objects it is trained on. Please note that our method uses 60 views for training and 6 views for inference. A-SDF* indicates A-SDF results compared to the geometry it trains on, while all other results use a sampling of the original geometry as ground truth. Lower score is better.}
\label{tab:sapien_rec}
\vspace*{-0.3cm}
\centering
\small
\resizebox{0.99\textwidth}{!}{
\begin{tabular}{lccccccccc}
\specialrule{.2em}{.1em}{.1em}
Method & Laptop & Stapler &Dishwasher &Eyeglasses &FridgeLR &FridgeUD &Drawer &DrawerUD& Drawer+Door\\
\hline
A-SDF* & 0.126 & 1.510 & 0.543 & \hspace*{-3.5pt}15.972 & 0.599 & 0.837 & 1.362 & 4.791 & 2.282 \\
A-SDF TTA* & 0.103 & 0.978 & 0.209 & 7.792 & 1.682 & 4.623 & 1.142 & 2.832 & 0.476 \\
\hdashline
A-SDF & 0.580 & 6.058 & 4.180 & \hspace*{-3.5pt}17.298 & 1.527 & 1.427 & 1.971 & 6.048 & 2.945 \\
A-SDF TTA & 0.542 & 5.358 & 3.756 & 9.052 & \textbf{1.351} & \textbf{0.842} & \textbf{1.689} & \textbf{4.139} & \textbf{1.082} \\
IDR 6 views & 1.656 & 1.113 & 4.139 & 1.386 & 7.915 & 1.826 & 4.202 & fails & \hspace*{-3.5pt}12.672 \\
IDR 60 views & \textbf{0.259} & 0.994 & \textbf{3.106} & \textbf{1.171} & \hspace*{-3.5pt}12.368 & 2.119 & 3.047 & 7.871 & \hspace*{-3.5pt}13.491 \\
Ours-ArtDef & 0.382 & 1.125 & 3.945 & 9.790 & 2.738 & 3.648 & 2.627 & 5.979 & 3.264 \\
Ours-ArtDef TTA & 0.355 & \textbf{0.936} & 3.936 & 7.894 & 2.063 & 3.649 & 2.745 & 5.912 & 3.243 \\
\specialrule{.1em}{.05em}{.05em}
\end{tabular}
}
\end{table*}
\begin{table*}[t]
\caption{\textbf{Comparison with IDR~\cite{yariv2020idr} for reconstructing unseen synthetic shapes (PSNR).} The training and inference procedures of IDR and our methods are the same as in Tab.~\ref{tab:sapien_rec}. Note that IDR trains a separate model for each articulation state of each object instance, it is trained on 60 views and then tested on 6 novel views of the same object,~\ie it trains and tests on the same instances, while our method is trained per category and is tested on unseen objects. IDR 60 views is trained on 60 views and tested on 6 novel views, offering an upper bound for the quality we can expect from our method. Higher score is better.}
\label{tab:sapien_psnr}
\vspace*{-0.3cm}
\addtolength{\tabcolsep}{-1pt}
\centering
\resizebox{0.99\textwidth}{!}{
\begin{tabular}{lcccccccccccc}
\specialrule{.2em}{.1em}{.1em}
Method  &Laptop &Stapler &Dishwasher &Eyeglasses &FridgeLR &FridgeUD  &Drawer&DrawerUD& Drawer+Door\\
\hline
IDR 6 views &13.32&9.75&17.64&11.49&10.81&13.47&15.26&fails&15.80\\
IDR 60 views & \textbf{20.70} & \textbf{22.40} & \textbf{24.10} & \textbf{26.59} & \textbf{20.01} &23.49&24.32&21.19&22.71\\
Ours-ArtDef &18.33&17.18&20.79&20.69&18.79&\textbf{23.72}& \textbf{24.65} &\textbf{22.25}&23.34\\
Ours-ArtDef TTA &17.84&18.15&20.87&20.89&18.94&23.52&24.27&22.20&\textbf{23.96} \\
\specialrule{.1em}{.05em}{.05em}
\end{tabular}
}
\end{table*}
\begin{figure*}
    \centering
    \includegraphics[width=1\linewidth]{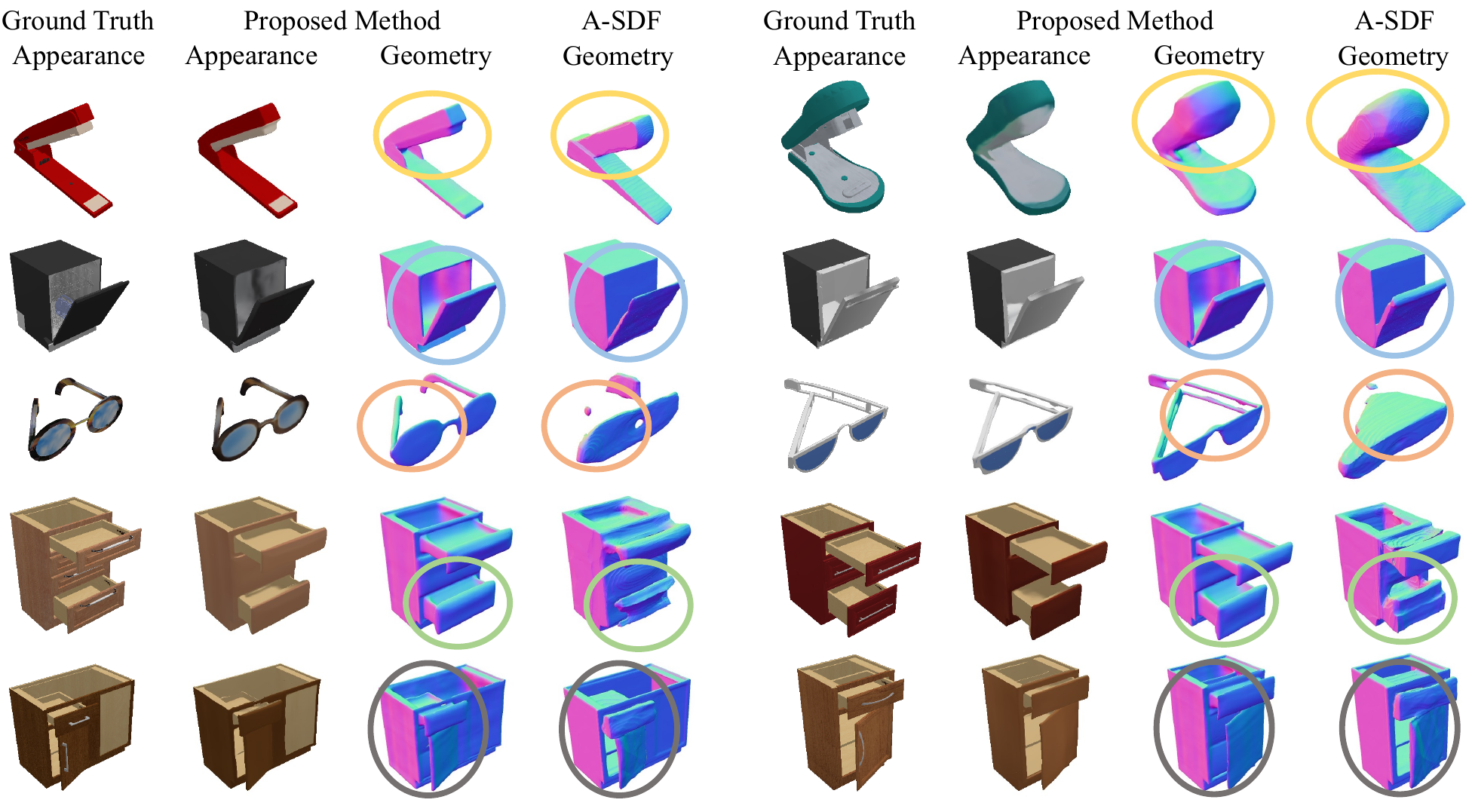}
    \vspace*{-0.7cm}
    \caption{\textbf{Reconstruction results from unseen data.} We compare to A-SDF on various joint types and combinations, including revolute, prismatic and multiple ones. While A-SDF trains using both geometry and articulation ground truth and only predicts geometry, our method also faithfully recovers appearance in addition to producing better geometry and more accurate articulations from only RGB supervision.}
    \vspace*{-0.2cm}
    \label{fig:sapien_rec}
\end{figure*}

To reconstruct unseen testing objects, we first optimize the codes (optionally with network weights) through backpropagation. Then we run another forward pass to extract the mesh and render images.
In Tab.~\ref{tab:sapien_rec} and Tab.~\ref{tab:sapien_psnr}, we compare both geometry and appearance with previous methods for 9 object categories, respectively.
We report the scores of our full model with articulation code sharing and using deformation field, with and without TTA. Please refer to the supplementary materials for a full list of results of each variant of our method.
For reference, we also compare with two IDR models.
Note that IDR does not have an embedding space to encode categorical priors and can only reconstruct a single static object or scene that it has been trained on,~\ie for IDR we train and test on the same object instances, one model per articulation state.
Since this is too computationally expensive, we randomly choose two objects with two articulations in each category and report the average PSNR as the PSNR on each category for IDR.
Our model is trained on 60 views per articulation and tested on 6 views per articulation for unseen objects, while IDR is trained on 60 or 6 views for one static object and tested on novel views of the same object with the same articulation.
Therefore, IDR trained on 60 views gives us a sense of the upper bound performance of our differentiable rendering system.

Note that meshes in the dataset are not watertight, so A-SDF processes them with the Manifold~\cite{manifoldsoftware} software in order to be able to sample SDF values, which does a resampling that leads to a thickening of the original meshes. 
In A-SDF~\cite{mu2021asdf} output geometry is evaluated against those thickened meshes that were used for training, rather than the original geometry, so we follow the same evaluation protocol and report it as A-SDF* in Tab.~\ref{tab:sapien_rec}, achieving comparable numbers to the original paper. 
All other method variants are evaluated with respect to a sampling of the original geometry, which may sometimes contain points sampled from hidden, internal structures, such as the bottom of a laptop's trackpad, but the fraction of these points is small.

We can see that despite being used on a harder task, our model performs on par with 60-view IDR on many categories such as furniture.
On other categories, our PSNR is significantly lower.
This is, because these categories tend to have higher frequency texture.
To show the benefit of the learned categorical prior introduced in our model, we further compare with an IDR model trained on 6 views.
Eventually, we would like to have a model that works nicely on unseen objects with only very limited observations.
This brings up the question: should we model the object by overfitting an IDR model with very few views or leveraging categorical priors---which is better?
The comparison between our method and 6-view IDR in Tab.~\ref{tab:sapien_psnr} and Tab.~\ref{tab:sapien_rec} clearly shows that by introducing a shape, articulation, and appearance prior, our model during inference significantly improves over an overfitted IDR with same number of views. We observe through visualization that our variants with deformation field performs better than without it when articulations involve topology change (\eg, a closed laptop being opened). And sometimes TTA may result in overfitting to appearance and makes the geometry prediction worse. 

We visualize the testing results from unseen views in Fig.~\ref{fig:sapien_rec} and compare with A-SDF (with TTA).
All models are trained with shared articulations and after testing time optimization.
The first two rows are single rotational joints followed by three rows of multi-joint categories with various joint type combinations.
We can see that for classes with small intra-class geometry variation such as laptop and dishwasher, A-SDF works well, which is also consistent with observations from Tab.~\ref{tab:sapien_rec}.
However, for classes with larger intra-class shape variations such as staplers and eyeglasses, A-SDF fails to capture the geometry details.
For example, for A-SDF, the geometry of the bottom part of both staplers are not correct, despite being directly optimized with 3D geometry during inference.
While A-SDF originally only shows results on rotational joints, we further test both methods on other joint types.
We can see that the performance of A-SDF on sliding joints is much worse than on rotational joints.
We reckon one reason is that A-SDF requires articulation ground truth input during training.
However, for drawers with different lengths, it is hard to define a single value shared across all objects.
In contrast, our method does not require articulation annotation and learns the articulation code through training. 
Please refer to the supplementary material for more results.

\subsection{Analysis}

\begin{table*}[t]
\caption{\textbf{Interpolation Chamfer-L1 error.} Evaluation details are the same as in Tab.~\ref{tab:sapien_rec}. Lower score is better.}
\label{tab:interpolation}
\vspace*{-0.3cm}
\centering
\small
\resizebox{0.98\textwidth}{!}{
\begin{tabular}{lcccccccccc}
\specialrule{.2em}{.1em}{.1em}
Method &Laptop &Stapler &Dishwasher &Eyeglasses &FridgeLR &FridgeUD &Drawer &DrawerUD& Drawer+Door\\
\hline
A-SDF* & 0.359 & 4.989 & 2.101 & \hspace*{-3.5pt}45.326 & 1.445 & 1.677 & 1.634 & 6.067 & 4.454 \\
\hdashline
A-SDF & 0.610 & 5.918 & 4.744 & \hspace*{-3.5pt}43.708 & \textbf{1.708} & \textbf{1.881} & \textbf{1.958} & 7.105 & 5.032 \\
Ours-base & 0.347 & \textbf{1.486} & 3.029 & 2.632 & 3.068 & 4.723 & 2.952 & 5.195 & \textbf{3.226} \\
Ours-Art & \textbf{0.309} & 1.716 & \textbf{2.807} & \textbf{2.588} & 3.860 & 3.357 & 3.086 & \textbf{4.151} & 3.845 \\
\specialrule{.1em}{.05em}{.05em}
\end{tabular}
\vspace*{-1.2cm}
}
\end{table*}

\begin{figure}
    \centering
    \includegraphics[width=1.0\linewidth,trim={0cm 0cm 0cm 0cm},clip]{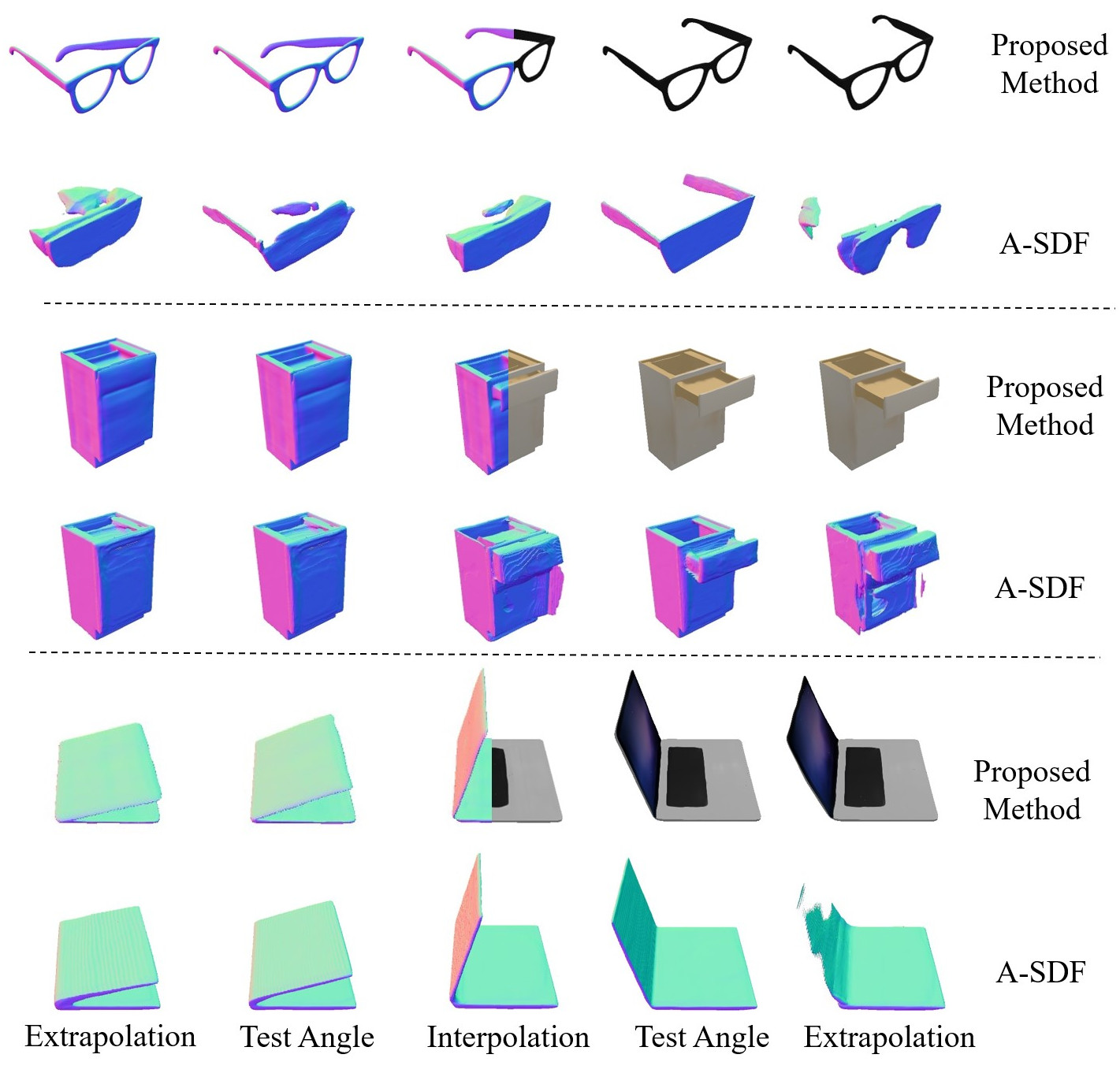}
    \caption{\textbf{Interpolation and extrapolation comparison.} For these unseen objects, we first infer the geometry, appearance and articulation codes for multiple testing states, then we interpolate / extrapolate the inferred codes to generate new articulations. The proposed method successfully generates shapes with articulations beyond the range seen during training for various joint types.}
    \label{fig:interpolation}
\end{figure}
\noindent\textbf{Interpolation and Extrapolation}
One application of the proposed method is to generate new articulations of an object through interpolation and extrapolation, given only a few training articulations.
To do so, we render a few images for two articulations, and optimize their shape, articulation, and appearance codes jointly with the images. 
After the two sets of codes are estimated, we interpolate/extrapolate a set of shape, articulation and appearance codes.
This procedure follows what is done in A-SDF~\cite{mu2021asdf}, to which we compare in Fig.~\ref{fig:interpolation} for different joint types and combinations. 
We run the two fastest variants of our method for this,~\ie those without deformation field.
Note that there is no TTA for either method as the two estimated code sets which are being interpolated need to share the same network weights. 
We see that A-SDF may fail on extrapolations more than 6\textdegree~ out of the training range for laptop and extrapolations for other classes, while our method is able to recover plausible shapes.
The quantitative results in Tab.~\ref{tab:interpolation} demonstrate that our interpolated models are geometrically more accurate than those of A-SDF for the majority of object categories. 

\begin{figure}
    \centering
    \includegraphics[width=0.8\linewidth]{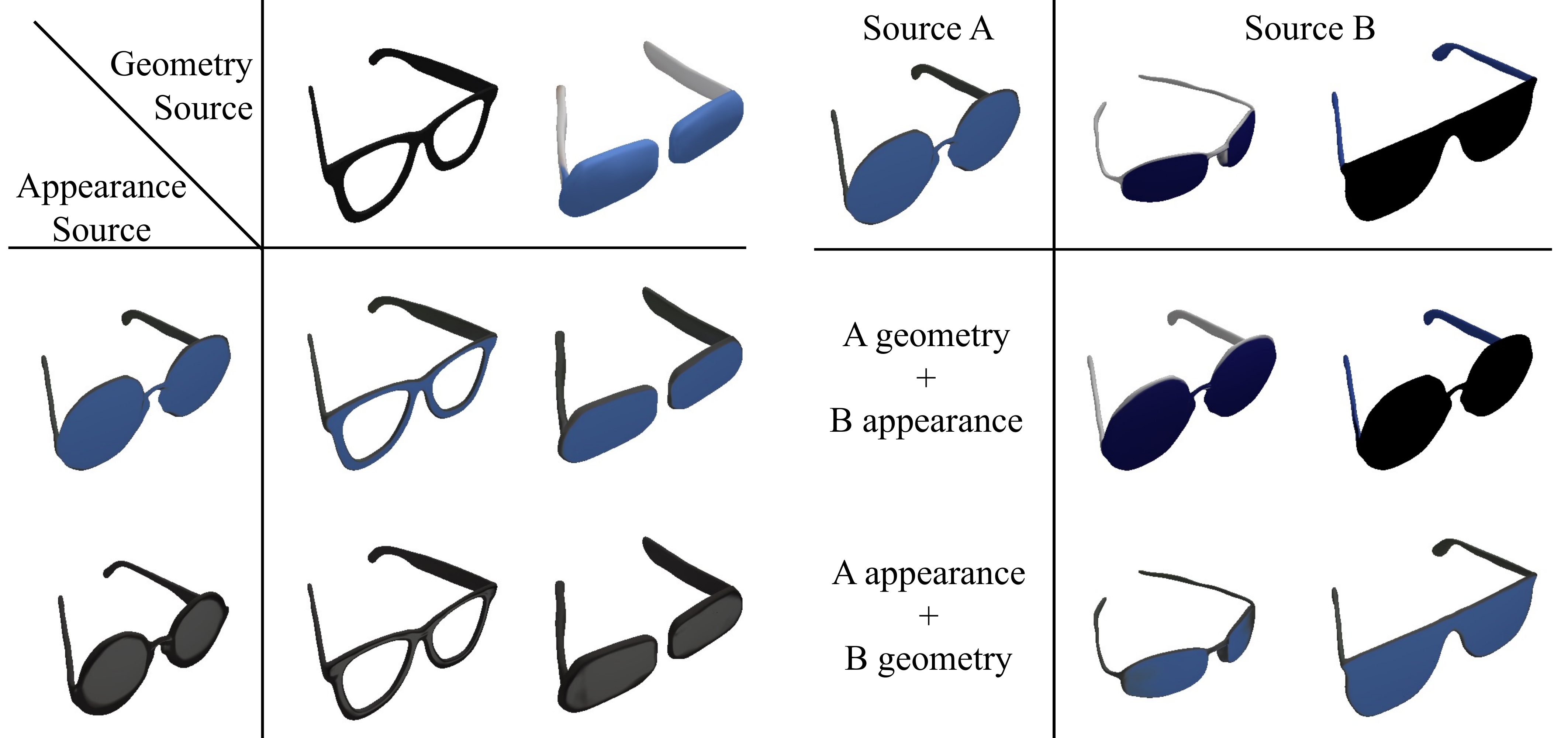}
    \caption{Disentangling shape and appearance for shape synthesis.}
    \label{fig:disentangle}
    \vspace*{-0.4cm}
\end{figure}
\noindent\textbf{Disentangling Geometry and Appearance.}
The separate geometry and appearance networks strongly enforce disentanglement, as inherited from IDR~\cite{yariv2020idr}.
One application of our method is to switch the geometry and appearance codes between different objects.
As shown in Fig.~\ref{fig:disentangle}, by replacing the appearance code with one from another pair of eyeglasses, we can faithfully create a new pair of eyeglasses that has a new appearance but the same geometry.
Note how the colors from the thin frame correctly map onto the frame of the new eyeglasses, despite the large geometry difference between the shapes.
Our learned embedding space that encodes categorical prior also helps with the correct mapping.

\subsection{Testing on Real-world RGB Images}

Since our method only uses images for supervision and does not require any 3D annotations that may be expensive, it can be easily set up to test on real data.
In this section, we directly test our proposed method (trained from synthetic data) on images captured in the real world.
We use a personal cell phone to record a static opened laptop or drawer with fixed focal length and exposure.
We then run Structure-from-Motion (SfM) algorithm~\cite{schoenberger2016sfm} on the captured frames to estimate the camera calibration parameters and their poses.
For each view, we then run https://remove.bg to estimate a segmentation mask for the foreground object.
The views shown in Fig.~\ref{fig:teaser} are the input images to our model. 
We test our model trained on synthetic data from SAPIEN with deformation field and shared articulation code on these real-world images. 
The shape, articulation, and appearance codes are initialized as described earlier, we then jointly fine-tune both network weights and codes on these images for 2000 iterations. 
At this point, we are able to reconstruct the static real objects. 
Then by replacing the inferred articulation code with the articulation codes learned during training, we are able to articulate the static reconstruction realistically. 

\subsection{Limitations}

While we push the boundaries for articulated shape reconstruction by removing limitations on required data and supervision compared to previous methods, remaining limitations exist.
Even though we are able to fine-tune our object models on real data, the domain gap from synthetic to real remains large, and the appearance prior we learn from the limited synthetic data is not powerful enough to explain general object appearance.
As a consequence, we make use of foreground masks to alleviate this problem.
A more elegant direction for future work could be to follow the example of VolSDF~\cite{volsdf:arxiv21} and extend the method towards unsupervised disentanglement of shape and appearance.
Another limitation is the current scaling behavior w.r.t. the number of joints: for modeling objects with $n$ joints with $m$ states, we need $m^n$ combinations.
This is only feasible for objects with a low number of joints.
Better joint priors and decoupling are interesting directions for future research.

\section{Conclusion}
\label{sec:conclusion}
In this paper, we set out to answer the research question whether it is possible to jointly learn a prior over the 3D geometry, articulation, and appearance of an entire class of objects, solely from photometric 2D observations, with no articulation annotations.
Our results show that this is not only possible but, given enough computational power, can be achieved with high fidelity and with remarkably little data.
In our experiments, we successfully fine-tune our model to real-world data using only 6 views and create animatable objects that closely resemble the real-world objects under articulations, interpolated, and extrapolated.
At the same time, our method is the first that not only handles revolute, but also prismatic joints and combinations thereof.
We hope that this encouraging result inspires further research into general-purpose object reconstruction.

\noindent\textbf{Acknowledgement} 
\scalebox{.95}[1.0]{We would like to thank Michael Goesele,} \scalebox{.97}[1.0]{Eddy Ilg, Zhaoyang Lv, Jisan Mahmud, Tanner Schmidt, and} \scalebox{.97}[1.0]{Anh Thai for helpful discussions.}
{\small
\bibliographystyle{ieee_fullname}
\bibliography{references}
}

\clearpage
\appendix
\section{Supplementary Material}

In this supplementary material, we describe details of dataset preparation in Sec.~\ref{supp_dataset} and implementation details for training, inference, and experiments on real data in Sec.~\ref{supp_implementation}. In Sec.~\ref{supp_results}, we provide more quantitative and qualitative results.
\subsection{Dataset}\label{supp_dataset}
\begin{figure}[t]
    \centering
    \includegraphics[width=\linewidth]{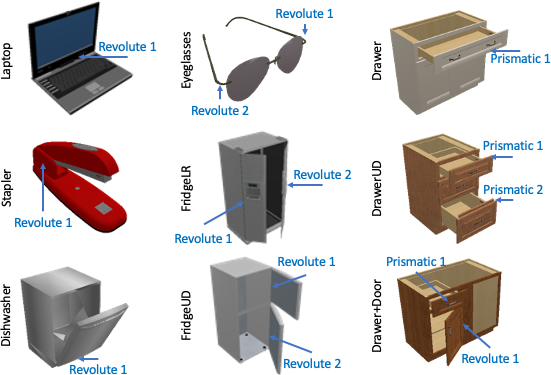}
    \caption{\textbf{Objects.} We show one sample object from each category in one articulation state. The joints and their types are annotated.}
    \label{fig:sapien}
\end{figure}

All experiments use SAPIEN~\cite{xiang2020sapien}, a large-scale, public domain dataset containing 2346 articulated objects across 46 categories.
We select six categories with representative articulation types and a sufficient number of instances: laptop, stapler, dishwasher, two-door fridge (LR for left and right, UD for up and down), eyeglasses, and storage furniture with drawer(s) (and door) (Drawer is for single-drawer furniture, DrawerUD for two-drawer furniture, and Drawer+Door for furniture that has one drawer and one door).
Note that we follow the same classification practice of A-SDF. For example, FridgeLR and FridgeUD both belong to the category of two-door fridges, but we still trained two separate models because A-SDF treated these two as two categories. We didn't try training on a combined category, but we expect it to work.
The different combinations of joint types and numbers in total make nine different categories.
We display one example of each category in Fig.~\ref{fig:sapien}.

\begin{table}[t]
\caption{\textbf{Dataset details.} We list the details of the SAPIEN data set for synthetic experiments. It covers a wide range of object classes and joint types. For each category, we show the number of joints of each type (revolute or prismatic), the number of object instances in the training and testing splits, the number of articulations sampled for training, and the number of views used for training and testing.}
\label{tab:dataset}
\addtolength{\tabcolsep}{-1pt}
\centering
\resizebox{0.5\textwidth}{!}{
\begin{tabular}{lccccccc}
\specialrule{.2em}{.1em}{.1em}
Category & \#joint & train / test split & \#art. & train/test \#view \\
\hline
Laptop & 1 revolute & 35/11 & 10 & 60/6 \\
Stapler & 1 revolute & 15/5 & 10 & 60/6 \\
Dishwasher & 1 revolute & 18/6 & 10 & 60/6 \\
Eyeglasses & 2 revolute & 48/14 & 36 & 60/6 \\
FridgeLR & 2 revolute & 8/3 & 36 & 60/6 \\
FridgeUD & 2 revolute & 12/4 & 36 & 60/6 \\
Drawer & 1 prismatic & 21/7 & 10 & 60/6 \\
DrawerUD & 2 prismatic & 27/9 & 36 & 60/6 \\
Drawer+Door & 1 revolute, 1 prismatic & 9/4 & 100 & 60/6 \\
\specialrule{.1em}{.05em}{.05em}
\end{tabular}
}
\end{table}

To render the shapes, we normalize them to fit in a unit sphere and make sure that the same object with different articulations are normalized in the same way (their non-motion parts are aligned). 
We use the SAPIEN simulation environment~\cite{xiang2020sapien} to render RGB images and corresponding masks.
During training and testing, we sample every 10\textdegree~ for rotational joints and 10 states in total for sliding joints. For multiple joints, we take all combinations of every single joint sampling. For each articulation, 60 views are sampled for training and 6 views for inference. 
Cameras are placed on vertices of a randomly rotated rhombicosidodecahedron for 60 views (octahedron for 6 views) with the object in its center.
The RGB images and masks are of resolution $640\times 480$.
These details are summarized in Tab.~\ref{tab:dataset}.

We set the angle range to train and test on revolute joints following A-SDF~\cite{mu2021asdf}. To evaluate interpolation, for every two neighboring testing articulations, we use the codes for these two articulations to interpolate the middle point. Concretely, for the stapler and dishwasher with training and testing angles \{0, 10, 20, 30, 40, 50, 60, 70, 80, 90\}, the angles used for evaluating interpolation are \{5, 15, 25, 35, 45, 55, 65, 75, 85\}.  For laptop,
angles used for training and testing are \{-72, -62, -52, -42, -32, -22, -12, -2, 8, 18\} and used for interpolation are \{-67, -57, -47, -37, -27, -17, -7, 3, 13\}. For the eyeglasses and fridge (fridgeLR and fridgeUD) with training and testing angles \{0, 10, 20, 30, 40, 50\} for each joint, the angles used for evaluating interpolation are \{5, 15, 25, 35, 45\}. For the drawer in storage furniture (Drawer, DrawerUD, Drawer$+$Door), we sample 10 articulations with equal distance for training and testing and use the 9 midpoints of the 10 articulations for interpolation. For the door in storage furniture (Drawer+Door), we use \{0, 10, 20, 30, 40, 50, 60, 70, 80, 90\} for training and testing, and \{5, 15, 25, 35, 45, 55, 65, 75, 85\} to evaluate interpolation.

\begin{figure}[t]
  \centering
  \includegraphics[width=1\linewidth]{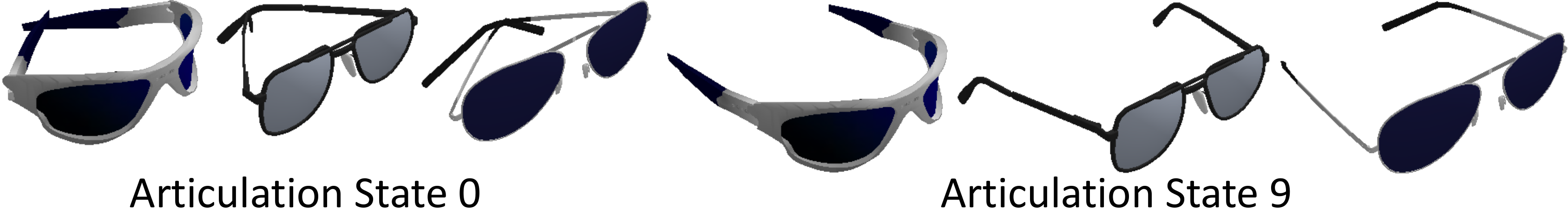}
  \caption{Articulations used in the experiments are not aligned.}
  \label{fig:align_articulation}
\end{figure}

To clarify, aligned articulation is \emph{not} required in training. In fact, SAPIEN objects are not aligned and we do not align them in our experiments. 
In Fig.~\ref{fig:align_articulation}, for example, eyeglasses at the articulation states 0 and 9 differ in the lens-leg angles.
However, if articulations are roughly aligned, we can also leverage it by sharing articulations (main paper Sec. 3.2) in variants Ours-Art/ArtDef.

\subsection{Implementation Details}\label{supp_implementation}

\subsubsection{Network Architecture}
The architecture of the geometry and appearance networks in our method follows exactly the description in IDR~\cite{yariv2020idr}. Concretely, the geometry network takes a 256-dimensional geometry feature and 3-dimensional 3D query location (and optionally a 8-dimensional articulation feature if running without deformation field) as input and predicts a single SDF value. When included, the deformation module takes in a 256-dimensional geometry feature, 3-dimensional 3D query location and a 8-dimensional articulation feature  as input and predicts a 3-dimensional displacement for the query point, which is added to the original query point and then passed to the geometry network. Both the geometry and the deformation network have eight fully connected hidden layers with a width 512 and a last fully connected layer with output dimension 1 or 3 for their corresponding predictions. There is a single skip connection from the input to the middle layer. The fully connected layers are interlaced with softplus activation in both networks. We follow the non-linear maps~\cite{yariv2020idr} on the input query points. We initialize the weights of the geometry network so that it produces an approximate SDF of a
unit sphere. 

In the appearance network, there are four fully connected layers with output dimension 512 and a last fully connected layer with output dimension 3 for color prediction. The input is a concatenation of the following: a 256-dimensional appearance feature for each object, a 3D surface point and its normal, and the viewing direction. We use the ReLU activation between hidden layers of the appearance network and tanh for the output to get valid color values.

\subsubsection{Training and Inference}
For training, latent codes are randomly initialized with $\mathcal{N}(0,\frac{1}{l})$, where $l$ is the code length.
We set $\rho=100, \lambda=0.1, \beta=0.0001$ for the loss in Eq. 7 of the main paper.
We start with $\alpha=50$ and multiply it by a factor of 2 every 50,000 iterations (up to a total of 5 multiplications).
The networks are trained using ADAM optimizer with a learning rate starting from 0.0001 and decreasing by a factor of 2 at the 50\% and 75\% point of the total number of iterations.

During inference, articulation codes are initialized to the mean of all learned articulation codes, while other codes are initialized as in training. To reconstruct unseen testing objects, we first optimize the geometry, articulation, and appearance codes through backpropagation for 600 iterations with learning rate starting from 0.009 and decreasing by a factor of 2 at 300 and 450 iterations. If we do test-time adaptation~\cite{mu2021asdf}, we further optimize both the codes and the network weights for another 600 iterations with learning rate  starting from 0.00005 and decreasing by a factor of 2 at 300 and 450 iterations. For both optimization stages, we start with $\alpha=50$ and multiply it by a factor of 2 every 100 iterations (up to a total of 5 multiplications). Then we run another forward pass to predict SDF values and render images.
The Marching Cubes algorithm is used to extract an approximate iso-surface given the predicted SDF values.

\subsubsection{Real Experiment Setup}
We test the model trained on synthetic laptops and drawers and directly apply the trained models to real-world phone-captured static objects.
We use a personal cell phone to record a static opened laptop or drawer with fixed focal length and exposure.
We then run Structure-from-Motion (SfM) algorithm~\cite{schoenberger2016sfm} on the captured frames to estimate the camera calibration parameters and their poses.
For each view, we then run https://remove.bg to estimate a segmentation mask for the foreground object.
We use seven input images to reconstruct the laptop and 24 images to reconstruct the drawer in Fig. 1 of the main paper. 
We test our model trained on synthetic data from SAPIEN with deformation field and shared articulation code on these real-world images. 
The shape, articulation, and appearance codes are initialized as described earlier, we then jointly fine-tune both the network weights and the codes on these images for 2000 iterations. 
At this point, we are able to reconstruct the static real objects. 
Then by replacing the inferred articulation code with the articulation codes learned during training, we are able to articulate the static reconstruction realistically. 

\begin{table*}[t]
\caption{\textbf{Reconstruction results on unseen synthetic shapes (Chamfer-L1).} We compare all variants of our proposed method. This table corresponds to Table 2 from the main paper.}
\label{tab:sapien_rec_methodvariants}
\centering
\small
\resizebox{0.99\textwidth}{!}{
\begin{tabular}{lccccccccc}
\specialrule{.2em}{.1em}{.1em}
Method & Laptop & Stapler &Dishwasher &Eyeglasses &FridgeLR &FridgeUD &Drawer &DrawerUD& Drawer+Door\\
\hline
Ours-base & 0.383 & 1.453 & 3.269 & 1.771 & 2.969 & 4.683 & 2.924 & 5.326 & \textbf{2.786} \\
Ours-Art & \textbf{0.328} & 1.560 & 2.962 & 1.735 & 3.955 & 3.332 & 3.114 & 4.185 & 3.416 \\
Ours-Def & 0.363 & 1.026 & 4.046 & 2.558 & 1.976 & 5.007 & 3.005 & 5.726 & 3.394 \\
Ours-ArtDef & 0.382 & 1.125 & 3.945 & 9.790 & 2.738 & 3.648 & \textbf{2.627} & 5.979 & 3.264 \\
Ours-base TTA & 0.345 & 1.336 & 3.187 & \textbf{1.606} & \textbf{1.637} & 4.614 & 2.940 & 5.100 & 2.899 \\
Ours-Art TTA & 0.475 & 1.400 & \textbf{2.881} & 1.659 & 2.635 & \textbf{3.238} & 3.135 & \textbf{4.166} & 3.897 \\
Ours-Def TTA & 0.333 & \textbf{0.815} & 4.046 & 2.026 & 2.244 & 4.669 & 3.042 & 5.335 & 3.652 \\
Ours-ArtDef TTA & 0.355 & 0.936 & 3.936 & 7.894 & 2.063 & 3.649 & 2.745 & 5.912 & 3.243 \\
\specialrule{.1em}{.05em}{.05em}
\end{tabular}
}
\end{table*}

\subsection{Results}\label{supp_results}

\begin{figure}[t]
  \centering
  \includegraphics[width=1\linewidth]{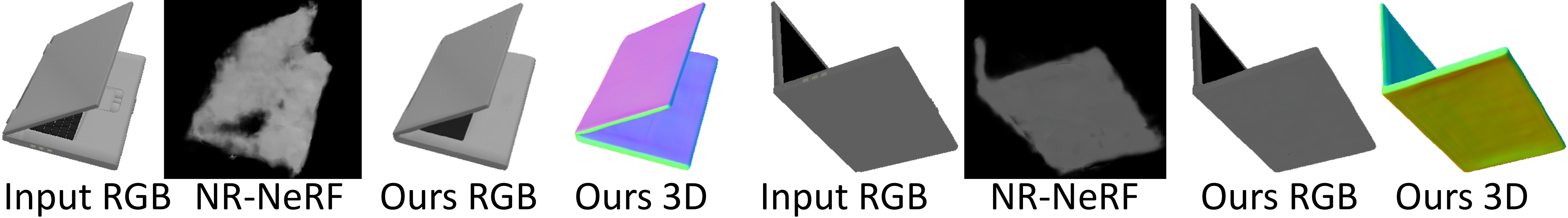}
  \caption{Comparison with NR-NeRF~\cite{tretschk2021nonrigid} on articulating a laptop.}
  \label{fig:nrnerf}
\end{figure}

\begin{figure}[t]
  \centering
  \includegraphics[width=1\linewidth]{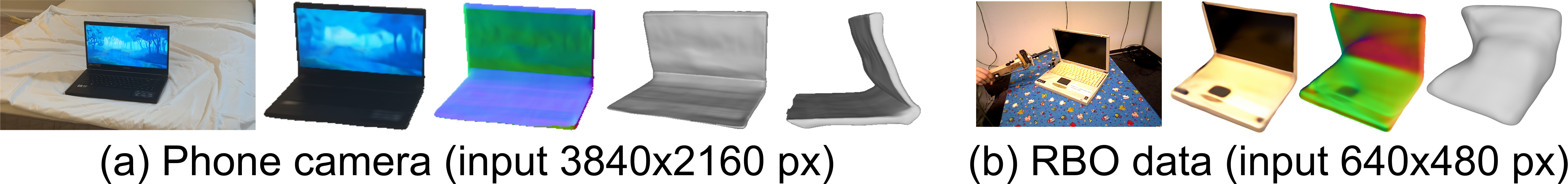}
  \caption{\textbf{Reconstruction from a single RGB image.} We show \mbox{input RGB, output appearance and normals, other untextured views}.}
  \label{fig:single_view}
\end{figure}

\paragraph{Full quantitative results.} In Tab.~\ref{tab:sapien_rec_methodvariants}, we show the full list of results for each variant of our method.
We observe that with deformation field it manages better with topology changes, but it takes longer to train. 
This is why the results of Ours-Def and Ours-ArtDef might be numerically worse as those models did not get to the same number of iterations in the same training time as without deformation field.
We also observe that sometimes TTA may cause the model to optimize towards a local minimum, \eg overfitting to appearance while making the geometry worse. The errors are larger on bulky objects like fridges, drawers, dishwashers, where the concave geometry is visible from very few views, so a method that only uses RGB information may not have enough coverage to carve the space out. While the numbers may not reflect all variants' strengths, combined with visualizations, we observe variants with deformation handle large topology changes better.
This is confirmed in Tab.~\ref{tab:sapien_rec_methodvariants} where ours-Def TTA performs the best on the stapler.

\paragraph{Comparison with NeRF-extension.} In Fig.~\ref{fig:nrnerf}, we show a comparison with NR-NeRF~\cite{tretschk2021nonrigid}, a representative NeRF extension to multi-view dynamic scenes. We ran its official code and our method on a SAPIEN laptop with the same 60 views $\times$ 10 angles setting. 
We observe that despite only recovering a single scene, NR-NeRF performs poorly due to large inter-frame movements.

\paragraph{Results on RBO dataset~\cite{rbo_dataset}.}
RBO dataset~\cite{rbo_dataset} only has monocular videos of articulated objects with fixed camera-object pose, so it is improper to evaluate our multi-view method. We still tested our single-view reconstruction on our real phone camera data and RBO in Fig.~\ref{fig:single_view}. It succeeds on high-res phone images, but on noisy, low-res RBO data it recovers plausible appearance but poor geometry that looks correct only from the input view. This strongly indicates that a few more views will be sufficient to disambiguate even noisy input. 
The Chamfer-L1 distance of the RBO example is 5.75 after scale determination, which is close to the DeepSDF error reported in A-SDF~\cite{mu2021asdf}, even though we do not use 3D input.

\subsection{Video}
Please refer to the video for more results on reconstruction, interpolation and extrapolation on testing synthetic data, as well as reconstruction and animation on real data.

\end{document}